\documentclass{article}

 \usepackage[preprint]{neurips_2026}

\usepackage[utf8]{inputenc} 
\usepackage[T1]{fontenc}    
\usepackage{xurl}
\usepackage{booktabs}       
\usepackage{amsfonts}       
\usepackage{nicefrac}       
\usepackage{microtype}      
\usepackage{xcolor}         

\usepackage{amsmath}
\usepackage{graphicx} 
\usepackage{multirow}
\usepackage{multicol}
\usepackage{wrapfig}
\definecolor{citecolor}{HTML}{2980b9}
\definecolor{linkcolor}{HTML}{c0392b}
\usepackage[hidelinks,breaklinks=true,colorlinks,citecolor=citecolor,linkcolor=linkcolor]{hyperref} 
\usepackage{cleveref}
\usepackage{caption}
\usepackage{amssymb}
\usepackage{graphicx}
\usepackage{booktabs}
\usepackage{pifont}
\usepackage{makecell}
\usepackage[table]{xcolor}

\newcommand{\best}[1]{\textbf{#1}}
\newcommand{\second}[1]{\underline{#1}}

\usepackage{tikz}
\usetikzlibrary{positioning, fit, arrows.meta, calc}
\usepackage{float}
\usepackage{subcaption}
\usepackage{algorithm}
\usepackage{algpseudocode}
\usepackage{amsmath}

\title{ROVER: Routing Object-Centric Visual Evidence for Grounded Multi-Image Reasoning}

\author{Guannan Lv, Ren Nie, Hongjian Dou, Tingting Gao \vspace{0.3cm}\\
  Kuaishou Technology\vspace{0.1cm}\\
  \texttt{\{lvguannan, nieren, douhongjian\}@kuaishou.com}
}

\begin{document}

\maketitle

\begin{abstract}
    Multimodal Large Language Models (MLLMs) have increasingly localized and interleaved visual evidence for deliberative reasoning.
Grounding-based approaches typically focus on regions of interest (RoIs) by injecting cropped image patches or RoI-specific features into the reasoning context.
However, such designs can weaken holistic scene understanding and inter-object relations, while incurring decoding costs that scale with the number and size of RoIs.
Alternatively, adaptive visual feature selection often requires fine-grained supervision or complex heuristics.
To address these limitations, we propose \textbf{ROVER} (\textbf{R}outing \textbf{O}bject-centric \textbf{V}isual \textbf{E}vidence for grounded multi-image \textbf{R}easoning), a lightweight, learnable plugin for efficient global visual evidence routing.
Upon each object grounding prediction, ROVER injects a step-specific token triplet to synergistically: (i) aggregate the ongoing reasoning context, (ii) distill intra-image cues into a visual working space via object-centric differential attention, and (iii) route and integrate history-aware evidence across objects and images within this space for subsequent reasoning.
We integrate ROVER into Qwen2.5-VL-7B and develop an interleaved SFT-to-GRPO training pipeline. Strictly adhering to the original datasets and evaluation protocols, our method achieves the best performance on MM-GCoT (+4.8\% answer accuracy, +14.6\% grounding accuracy) and VideoEspresso (+8.6\% answer accuracy). The VideoEspresso-trained model demonstrates strong transferability, outperforming the base model by +4.7\% on average across diverse benchmarks.
\end{abstract}

\section{Introduction}
\label{sec:introduction}
Multimodal Large Language Models (MLLMs) have rapidly advanced in both vision--language understanding and generation, narrowing the long-standing divide between visual perception and language-driven reasoning~\cite{team2023gemini, liu2023llava, wang2024qwen2}. 
Recent research has increasingly equipped MLLMs with language-centric reasoning mechanisms, among which Chain-of-Thought (CoT) has emerged as a dominant paradigm.
These strategies have consistently yielded substantial gains across a broad spectrum of multimodal tasks~\cite{wei2022chain, kojima2022large, su2025thinking, zhang2023multimodal, lu2023mathvista}.

In vision--language reasoning, a widely adopted paradigm encodes images once and performs text-only CoT reasoning over fixed visual features~\cite{zhang2023multimodal, he2025mmboundaryadvancingmllmknowledge, shen2025satori, su2025thinking}. 
While effective for many tasks, this approach often falls short when fine-grained perception is required~\cite{shi2025eagle, tong2024cambrian, tong2024eyes}, as it prevents the model from explicitly revisiting and reorganizing visual evidence along the evolving reasoning trajectory.
Motivated by this limitation, recent work shifts from thinking \emph{about} images to thinking \emph{with} images~\cite{openai_o3_o4mini_2025, su2025thinking}, treating visual representations as revisitable intermediate states during generation. 
A representative advancement in this direction is visual interleaving, which dynamically injects targeted visual evidence into the reasoning stream~\cite{wang2025simpleo3, man2025argus, chen2025mint, hu2024visual}.

Despite their effectiveness, most interleaving designs remain region-centric.
They explicitly target regions of interest (RoIs) by injecting cropped image patches or RoI-specific visual features into the decoding stream \cite{wang2025simpleo3, shao2024visual, hu2024visual, yu2025introducing, wang2025vgrvisualgroundedreasoning, man2025argus}.
Such mechanisms face two structural limitations. First, focusing on isolated regions neglects the broader scene context and inter-object relations. Lacking sufficient holistic understanding, MLLMs tend to fall back on language priors and potentially induce hallucinations~\cite{chen2025ict}. Moreover, ignoring inter-object relations can lead to biased conclusions, particularly in human-in-the-loop scenarios where coarse and sparse user-provided region prompts often omit critical contextual cues~\cite{yang2023mm, zhao2024chatspot}.
Second, the decoding cost scales significantly with the spatial size of RoIs, making it expensive to process high-resolution and multi-object scenes.

Beyond RoI-based revisitation, adaptive visual feature selection or pruning typically aims to retain and interleave step-relevant visual features \cite{chen2025mint, gao2025interleaved, zhong2025focus}.
However, existing approaches are largely confined to single-image scenarios and necessitate expensive fine-grained annotations or intricate heuristics, hindering scalable multi-image reasoning.
Another line of work augments MLLMs with external tools or executable programs to iteratively acquire and manipulate evidence \cite{bai2025multi, gupta2023visual, hu2024visual, zheng2025deepeyes, li2026deepscan, qiao2025v, wang2025pixel}.
While promising, tool-augmented frameworks often incur high latency and complexity, with potential execution brittleness complicating seamless integration.

These limitations call for a decoding-efficient mechanism that transcends isolated regions to facilitate holistic scene understanding and the modeling of inter-object relations.
In this work, we propose \textbf{ROVER} (\textbf{R}outing \textbf{O}bject-centric \textbf{V}isual \textbf{E}vidence for grounded multi-image \textbf{R}easoning), which extends RoI-based revisitation into a \emph{structured visual evidence routing process co-evolving with language reasoning}. This synergistic architecture conceptually aligns with the dual-process cognitive framework~\cite{kahneman2011thinking}. 
As an intuitive frontend, ROVER rapidly aggregates object-centric visual context for the MLLM backend's deliberate reasoning. In multi-hop scenarios, the MLLM can dynamically initiate further grounding steps to accumulate supplementary visual evidence.

To operationalize this intuitive frontend, the routing process draws inspiration from human visual cognition---specifically selective attention~\cite{lewis2018removal}, working memory~\cite{oberauer2019working}, and return fixations~\cite{zhang2022look, nikolaev2025refixation}. 
Upon each object grounding prediction, the reasoning context is first absorbed into a compact representation. Serving as an active routing node, each grounded object then distills contextual cues while suppressing distractors to incorporate broader scene context, and archives this information into the history-aware Visual Working Space (VWS). Attending to the VWS resembles return fixations and iteratively integrates historical evidence across objects and images. In practice, ROVER encodes this mechanism into a constant-length token triplet, bridging local details and global reasoning while bypassing the overhead of injecting variable-length RoI features.

We integrate ROVER into Qwen2.5-VL-7B~\cite{bai2025qwen25vltechnicalreport} and develop a unified interleaved SFT-to-GRPO~\cite{shao2024deepseekmath} training pipeline.
Under strictly controlled settings relying solely on the original datasets, our method achieves the best performance on MM-GCoT~\cite{wu2025grounded} (+4.8\% answer accuracy, +14.6\% grounding accuracy), and improves answer accuracy by +8.6\% over the previous best on VideoEspresso~\cite{Han_2025_CVPR}.
Notably, despite being fine-tuned exclusively on VideoEspresso, the resulting model demonstrates robust transferability, outperforming the base model by +4.7\% on average across diverse benchmarks (e.g., +2.2\% on Mantis~\cite{Jiang2024MANTISIM}, +4.8\% on V-Star~\cite{wu2024v}, and +5.9\% on TreeBench~\cite{wang2025traceable}).

Our main contributions are summarized as follows:
\begin{itemize}
    \item We propose \textbf{ROVER} (\textbf{R}outing \textbf{O}bject-centric \textbf{V}isual \textbf{E}vidence for grounded multi-image \textbf{R}easoning), a lightweight learnable plugin that implements global object-centric visual evidence routing during chain-of-thought generation by appending a constant-length token triplet after each grounding prediction.
    \item We devise an object-centric differential attention mechanism to distill complementary cues while suppressing visual distractors, alongside a Visual Working Space (VWS) as a structured routing substrate. This architecture consolidates object-centric cues and enables history-aware routing of visual evidence across objects and images during decoding.
    \item We integrate ROVER into Qwen2.5-VL-7B and develop a unified interleaved SFT-to-GRPO training pipeline, yielding consistent gains on both single-image and multi-image grounded reasoning tasks in controlled settings, and superior generalization to diverse benchmarks.
\end{itemize}
\section{Related work}
\begin{figure}[t]
\centering
\includegraphics[width=\linewidth]{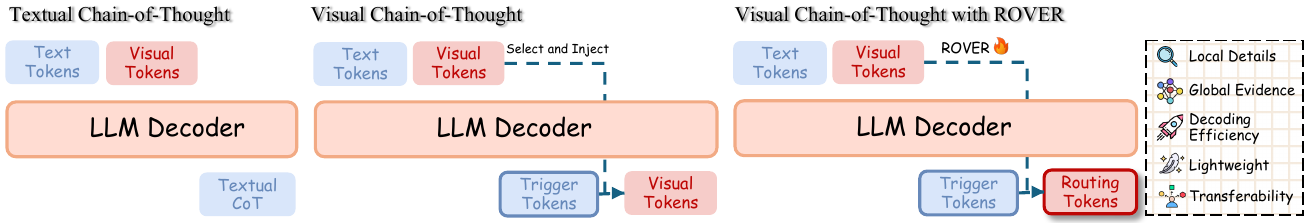}
\caption{\textbf{Comparison between our method and prior paradigms.}
Compared to textual CoT and existing visual CoT approaches, ROVER preserves local object details while seamlessly routing evidence across objects and images. It is learnable, lightweight, and decoding-efficient by injecting a constant-length token triplet per grounding step, and exhibits strong transferability.}
\label{fig:comparison}
\end{figure}

\begin{figure}[t]
\centering
\includegraphics[width=\linewidth]{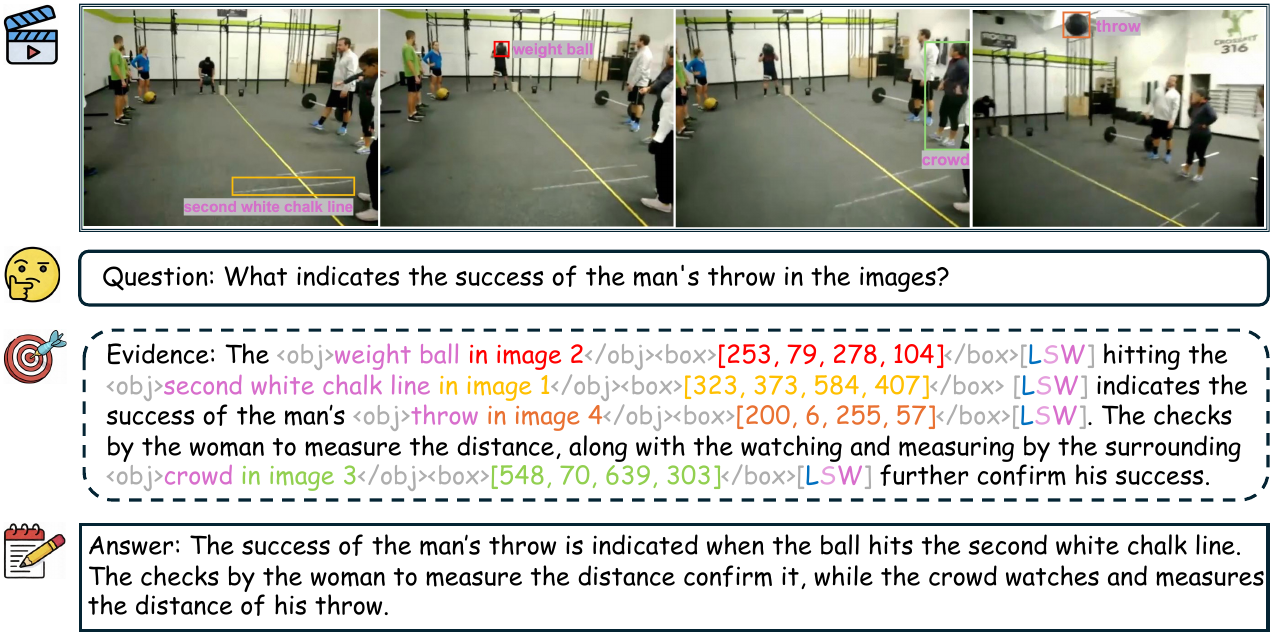}
\caption{\textbf{Qualitative example with ROVER token insertions.} Given a multi-image query from VideoEspresso~\cite{Han_2025_CVPR}, the model grounds key entities across images (ball in image~2, chalk line in image~1, throw in image~4, and crowd in image~3) and composes evidence to support the final answer. We mark the insertion positions of the \textbf{Link}/\textbf{Sift}/\textbf{Weave} triplet with [LSW] in the evidence. Bounding boxes are reported in absolute pixel coordinates as $[x_{\min}, y_{\min}, x_{\max}, y_{\max}]$.}
\label{fig:case}
\end{figure}

\paragraph{Reasoning with LLMs.}
Chain-of-thought (CoT) prompting improves LLM reasoning by eliciting explicit intermediate steps that guide the model toward the final solution~\cite{wei2022chain}.
Expanding upon this foundation, follow-up work has developed a variety of CoT-related paradigms, such as zero-shot CoT prompting~\cite{kojima2022large} and automated CoT synthesis~\cite{zhang2022automatic}.
Beyond prompting, recent studies leverage reinforcement learning to directly optimize outcome-driven objectives along CoT-style solution trajectories~\cite{schulman2017proximal, lightman2023let, shao2024deepseekmath}, leading to strong gains on difficult text-only benchmarks, including STEM problem solving and code generation~\cite{jain2024livecodebench, rein2024gpqa}.
Nevertheless, relying solely on textual CoT often proves fundamentally inadequate for complex vision-centric or vision--language tasks~\cite{fu2024blink, tong2024cambrian}, thereby necessitating multimodal approaches that can actively revisit visual evidence to substantiate intermediate reasoning steps.

\paragraph{Reasoning with MLLMs.}
Multimodal Large Language Models (MLLMs) have revolutionized open-ended visual understanding and complex reasoning tasks~\cite{liu2023llava}.
This rapid progress is propelled by innovations in visual instruction tuning, robust vision--language alignment, high-resolution perception mechanisms, and the curation of reasoning-intensive multimodal datasets~\cite{Bai2023QwenVLAF, man2025argus, wang2024qwen2, li2024llava-ov}. These foundational advancements also underpin the emergence of increasingly capable proprietary systems that demonstrate sophisticated reasoning in real-world scenarios~\cite{achiam2023gpt, team2023gemini, AnthropicModelCA}.

\paragraph{Visual Chain-of-Thought.}
As MLLMs mature, visual chain-of-thought (visual CoT) has emerged to ground intermediate reasoning in visual evidence.
This body of research encompasses RoI-centric revisitation, tool-augmented visual reasoning, and adaptive visual feature selection.
For example, Argus~\cite{man2025argus} revisits RoI-specific visual signals during reasoning, while Visual Sketchpad~\cite{hu2024visual} augments reasoning with tool-generated intermediate visual artifacts (e.g., drawing auxiliary lines) to support visual problem solving.
Existing adaptive methods, whether heuristic (e.g., ICoT~\cite{gao2025interleaved}) or learned (e.g., MINT-CoT~\cite{chen2025mint}), primarily target single-image reasoning.
This focus, coupled with reliance on sensitive hyperparameters or expensive annotations, limits their extension to multi-image reasoning scenarios.
Collectively, these trade-offs motivate context-preserving revisitation with global, object-centric visual evidence routing for multi-image reasoning.
\section{Method}
\label{sec:method}

We propose ROVER, a mechanism that equips autoregressive multimodal generation with an explicit, object-level pathway for global visual evidence routing.
In \Cref{sec:ROVER}, we detail the proposed framework (depicted in \Cref{fig:rover_overview} and compared to prior paradigms in \Cref{fig:comparison}). 
It comprises four key components: (i) a trigger-and-insert interface that determines when to initiate routing, (ii) the \textbf{Link}/\textbf{Sift}/\textbf{Weave} token triplet serving as routing primitives, (iii) object-centric differential cross-attention for evidence routing within each image, and (iv) the Visual Working Space (VWS) that enables history-aware visual evidence routing across objects and images. 
Finally, we present our training recipe in \Cref{sec:Training}, combining tailored interleaved SFT and GRPO.
\subsection{ROVER}
\label{sec:ROVER}

\begin{figure}[t]
\centering
\includegraphics[width=\linewidth]{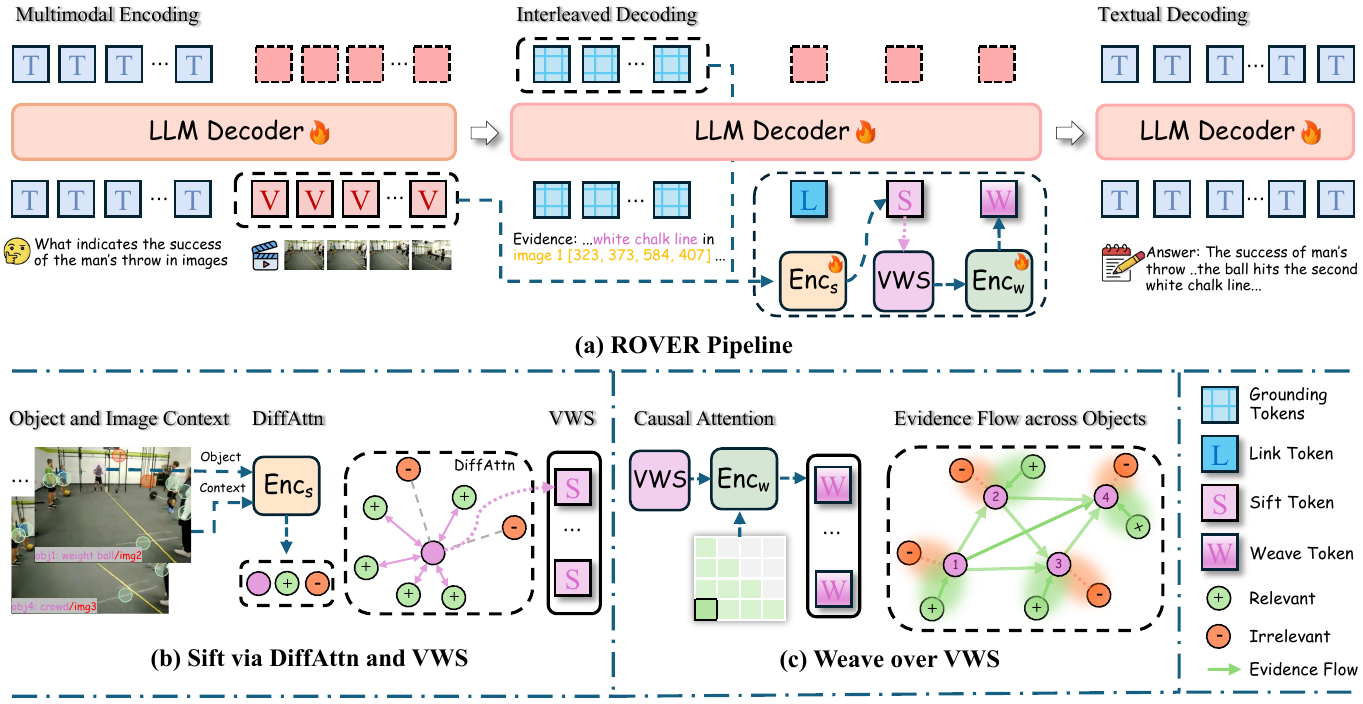}
\caption{\textbf{Overview of ROVER.} (a) \textbf{ROVER Pipeline.} Triggered by each object grounding, ROVER injects a \textbf{Link}/\textbf{Sift}/\textbf{Weave} token triplet to route visual evidence through the VWS. The MLLM then autoregressively interleaves reasoning and subsequent grounding steps, concluding with a pure-text answer. (b) \textbf{Sift via DiffAttn and VWS.} Leveraging object-centric differential attention, \textbf{Sift} distills complementary visual context while suppressing irrelevant regions to populate the VWS with compact object-centric cues. (c) \textbf{Weave over VWS.} Via history-aware attention, \textbf{Weave} routes and integrates relevant evidence from previously grounded objects into the current reasoning state.}
\label{fig:rover_overview}
\end{figure}

\paragraph{Triggering.}
\label{sec:ROVER_trigger}
ROVER initiates visual evidence routing upon detecting a valid event $o_k$. Formally, $o_k$ encapsulates an object phrase (any visually localizable concept, from tangible items to specific actions and visual states), its image index, and spatial localization (typically a bounding box). Triggers can originate from diverse sources, including intrinsic model-emitted grounding patterns~\cite{man2025argus}, user-provided region prompts (e.g., clicks or boxes)~\cite{zhao2024chatspot}, and external detector proposals~\cite{liu2024grounding}.

In this work, we primarily focus on the model-emitted grounding trigger: the routing event is activated upon the generation of a valid grounding pattern---e.g., <obj>the man in image 1</obj> <box>[$x_{\min}, y_{\min}, x_{\max}, y_{\max}$]</box> (see \Cref{fig:case}). Here, $[x_{\min},y_{\min}]$ and $[x_{\max},y_{\max}]$ denote the top-left and bottom-right corners of the predicted bounding box in image coordinates.

\paragraph{Routing Primitives.} 
After each routing event, ROVER appends an explicit object-level routing interface formatted as a constant-length token triplet---\textbf{Link}, \textbf{Sift}, and \textbf{Weave}. 
This is represented as $\mathbf{T}_k \in \mathbb{R}^{3\times d}$, where $d$ is the hidden size:
\begin{equation}
\mathbf{T}_k=\big[\mathbf{t}^{\mathrm{Link}}_k;\ \mathbf{t}^{\mathrm{Sift}}_k;\ \mathbf{t}^{\mathrm{Weave}}_k\big].
\end{equation}

As depicted in \Cref{fig:rover_overview}, the learnable \textbf{Link} embedding absorbs the evolving decoding context, compactly encapsulating the current reasoning state~\cite{gao2025interleaved}. \textbf{Sift} populates the VWS by distilling object-anchored intra-image context. For instance, when grounding a braking car on the street, it can attend to co-occurrent traffic lights or crosswalks to gather candidate visual cues and facilitate holistic scene understanding for subsequent reasoning. \textbf{Weave} then leverages the VWS to integrate historical evidence across objects and images. This design imposes a constant three-token overhead per grounded object, independent of its spatial size.

\paragraph{Routing Modules.}
Both \textbf{Sift} and \textbf{Weave} routing modules are instantiated as lightweight single-layer Transformer blocks~\cite{vaswani2017attention} comprising cross-attention and feed-forward networks (FFNs).
Let $\mathrm{Enc}(\mathbf{X};\mathbf{S},\mathbf{M})$ denote a generic block where the query $\mathbf{X}$ attends to the source $\mathbf{S}$ (serving as both key and value, i.e., $\mathbf{K}=\mathbf{V}=\mathbf{S}$) under mask $\mathbf{M}$.
Sharing this architectural backbone, $\mathrm{Enc}_{\mathrm{Sift}}$ adopts differential attention, whereas $\mathrm{Enc}_{\mathrm{Weave}}$ relies on standard attention.

\paragraph{Sift with DiffAttn.}
Given an input image $\mathcal{I}_m$, the vision encoder yields a sequence of patch tokens $\mathbf{V}_m\in\mathbb{R}^{N_m\times d}$.
For the $k$-th valid object bounding box $\mathbf{b}_k$ in image $\mathcal{I}_{m(k)}$, let $\Omega(\mathbf{b}_k)$ denote the indices of patches overlapping the box, and $\bar{\Omega}(\mathbf{b}_k)$ denote the remaining non-RoI patch indices.
Following~\cite{guo2024regiongpt, li2024llama}, we first compute a compact RoI query $\mathbf{q}_k$ via average pooling:
\begin{equation}
\mathbf{q}_k=\mathrm{AvgPool}\big(\mathbf{V}_{m(k)}[\Omega(\mathbf{b}_k)]\big)\in\mathbb{R}^{d}.
\end{equation}
As illustrated in \Cref{fig:rover_overview}, \textbf{Sift} is designed to extract complementary context while suppressing irrelevant distractors. We adopt differential attention~\cite{ye2024differential} as the core operator in $\mathrm{Enc}_{\mathrm{Sift}}$.
Formally, given query $Q$, key $K$, and value $V$, the operator computes:
\begin{equation}
\mathrm{DiffAttn}(Q,K,V)=\Big(\mathrm{softmax}(\tfrac{Q_1K_1^\top}{\sqrt{d_h}})-\lambda \cdot\mathrm{softmax}(\tfrac{Q_2K_2^\top}{\sqrt{d_h}})\Big)V,
\end{equation}
where query/key projections are split into two heads $[Q_1;Q_2]=QW_Q$, $[K_1;K_2]=KW_K$, and $\lambda$ is a learnable parameter initialized as in~\cite{ye2024differential}.
\textbf{Sift} executes object-centric differential attention to generate $\mathbf{t}^{\mathrm{Sift}}_k$, querying the complementary context $\mathbf{V}_{m(k)}[\bar{\Omega}(\mathbf{b}_k)]$ with the object query $\mathbf{q}_k$:
\begin{equation}
\mathbf{t}^{\mathrm{Sift}}_k=\mathrm{Enc}_{\mathrm{Sift}}\!\Big(\mathbf{q}_k;\ \mathbf{V}_{m(k)}[\bar{\Omega}(\mathbf{b}_k)],\ \mathbf{M}^{\mathrm{Sift}}_k\Big).
\end{equation}
With the residual connection~\cite{he2016deep} inherently retaining the object query $\mathbf{q}_k$, the attention mechanism is exclusively dedicated to routing complementary context from non-RoI regions (e.g., scene background or surrounding entities).
In the rare edge case where $|\bar{\Omega}(\mathbf{b}_k)| = 0$ (e.g., full-image box), differential comparison becomes inapplicable, and we default to $\mathbf{t}^{\mathrm{Sift}}_k=\mathbf{q}_k$.

\paragraph{Weave over VWS.}
We maintain the VWS as a unified, history-ordered routing substrate:
\begin{equation}
\mathcal{W}_k=[\mathcal{W}_{k-1};\mathbf{t}^{\mathrm{Sift}}_k]\in\mathbb{R}^{k\times d},\quad \mathcal{W}_0=\varnothing.
\end{equation}
Importantly, as shown in \Cref{fig:rover_overview}, this workspace is shared across all routing events irrespective of the source image index, thus allowing $\mathcal{W}_k$ to form a global memory of grounded objects across multiple images.
We compute $\mathbf{t}^{\mathrm{Weave}}_k$ with a single-layer encoder that routes and integrates visual evidence through the VWS, where $\mathbf{t}^{\mathrm{Sift}}_k$ attends to $\mathcal{W}_k$ under an autoregressive mask:
\begin{equation}
\mathbf{t}^{\mathrm{Weave}}_k=\mathrm{Enc}_{\mathrm{Weave}}\!\Big(\mathbf{t}^{\mathrm{Sift}}_k;\ \mathcal{W}_k,\ \mathbf{M}^{\mathrm{Weave}}_k\Big).
\end{equation}

\paragraph{Decoding.}
After the $k$-th routing event, we append $\mathbf{T}_k$ to the input embedding sequence, allowing the backbone LLM to perform a forward pass and then autoregressively interleave textual reasoning with further grounding steps.
For a trajectory with $K$ routing events, ROVER introduces exactly $3K$ additional tokens (i.e., $L_K=L_0+3K$), independent of the spatial size of RoIs.

\subsection{Training}
\label{sec:Training}

\paragraph{Overview.}
Featuring the interleaving of text tokens and ROVER triplets, our training pipeline proceeds in two stages: (i) interleaved supervised fine-tuning (SFT), which empowers the model to emit accurate grounding patterns and task-specific responses, and (ii) interleaved Group Relative Policy Optimization (GRPO)~\cite{shao2024deepseekmath} to optimize trajectory rewards for better answer accuracy.

\paragraph{Interleaved SFT.}
Given a dataset of multimodal instruction-response pairs, we construct training sequences that alternate natural language tokens with closed grounding patterns (see \Cref{fig:case}).
During teacher forcing, whenever a closed and valid grounding pattern is completed, we trigger a routing event and insert the \textbf{Link}/\textbf{Sift}/\textbf{Weave} token triplet $\mathbf{T}_k$ into the input embedding sequence.
We minimize the masked negative log-likelihood:
\begin{equation}
\mathcal{L}_{\mathrm{SFT}}
= \mathbb{E}_{t \mid m_t = 1}
\left[
- \log p_{\theta}\big(y_t \mid y_{<t}, \mathcal{I}\big)
\right].
\end{equation}
Here, the target tokens $y_t$ comprise both textual and grounding outputs, conditioned on the preceding context $y_{<t}$ and the visual input $\mathcal{I}$. We set $m_t=1$ for these targets and $m_t=0$ for the inserted token triplets.
Although these inserted tokens are not directly supervised, they affect the hidden states used to predict subsequent target tokens, and are thus optimized end-to-end via backpropagation driven by the language modeling objective.

\paragraph{Interleaved GRPO.}
Building upon the SFT checkpoint, we tailor GRPO~\cite{shao2024deepseekmath} to the interleaved multimodal setting to further optimize performance. For each prompt, we sample a group of $G$ trajectories $\{\tau_i\}_{i=1}^{G}$ from the old policy $p_{\theta_{\mathrm{old}}}$ and compute task-specific rewards $r(\tau_i)$. The rewards are then normalized to obtain advantages:
\begin{equation}
\hat{A}(\tau_i)=\frac{r(\tau_i)-\mu_r}{\sigma_r+\epsilon},
\end{equation}
where $\mu_r$ and $\sigma_r$ are the group-wise mean and standard deviation of $\{r(\tau_i)\}_{i=1}^{G}$. We use the following KL-regularized GRPO loss:
\begin{equation}
\mathcal{L}_{\mathrm{GRPO}}
= \mathbb{E}_{\tau_i \sim p_{\theta_{\mathrm{old}}},\, t \mid m_t = 1}
\left[
- \frac{p_{\theta}(y_t \mid y_{<t})}{p_{\theta_{\mathrm{old}}}(y_t \mid y_{<t})}\hat{A}(\tau_i)
+ \beta\,\mathrm{D}_{\mathrm{KL}}(p_{\theta} \| p_{\mathrm{ref}})
\right].
\end{equation}
Here, $\mathrm{D}_{\mathrm{KL}}(p_{\theta} \| p_{\mathrm{ref}})$ denotes the token-level KL divergence between the current policy $p_{\theta}$ and the reference policy $p_{\mathrm{ref}}$ (initialized from the SFT model), weighted by a coefficient $\beta$. Note that the ROVER triplet consists of deterministically inserted tokens; hence, we exclude them from the loss calculation by setting $m_t=0$.
\section{Experiments}
\label{sec:experiments}

This section evaluates ROVER across diverse grounded reasoning scenarios. We begin by detailing the implementation specifics and evaluation protocols in \Cref{sec:exp_setup}. Next, we present comparative results on (i) multi-image grounded reasoning with VideoEspresso (\Cref{sec:exp_videoespresso}) and (ii) single-image grounded reasoning on MM-GCoT (\Cref{sec:exp_mmgcot}), benchmarking against standard text-only grounded chain-of-thought (GCoT) approaches and additional vision-centric RoI-based revisitation strategies. Finally, in \Cref{sec:exp_transfer}, we assess backbone compatibility and robust zero-shot transfer capabilities across various held-out benchmarks.

\subsection{Experimental setup}
\label{sec:exp_setup}

\paragraph{Implementation Details.}
We build ROVER upon Qwen2.5-VL-7B~\cite{bai2025qwen25vltechnicalreport}, adhering to the official protocols of VideoEspresso~\cite{Han_2025_CVPR} and MM-GCoT~\cite{wu2025grounded}.
These benchmarks pose complementary challenges. VideoEspresso requires long-form, multi-step grounded reasoning across multiple images (e.g., video keyframes). MM-GCoT, in contrast, emphasizes GCoT generation from a single image, requiring precise inter-object reasoning.
For VideoEspresso, we use only interleaved SFT, since its open-ended nature precludes reliable RL rewards. For MM-GCoT, we apply the full SFT-to-GRPO pipeline under a controlled setting: we split the original training set between the SFT and RL stages, strictly avoiding any external data. During training, we fine-tune all parameters except the frozen vision encoder. Further details are provided in Appendix~\ref{sec:appendix_impl}.

\paragraph{Evaluation Protocols.}
In accordance with the official VideoEspresso protocol, we report category-wise scores and the overall average, supplemented by an LLM-as-a-judge evaluation for open-ended responses (see details in \Cref{sec:exp_videoespresso}).
For MM-GCoT, we report answer accuracy (A-Acc), grounding accuracy (G-Acc), and answer--grounding consistency (Consist.).
Additional evaluation metrics and prompt template details are provided in Appendix~\ref{sec:appendix_eval}.

\paragraph{Baselines and Variants.}
We structure our comparisons across two paradigms:
(1) Text-only GCoT: representative baselines on each dataset relying on static visual features.
(2) Visual CoT: vision-centric RoI-based revisitation strategies compared against our routing mechanism, following the taxonomy in Argus~\cite{man2025argus}:
(i) RoI-reencoding, which re-encodes cropped image patches (e.g., Visual CoT~\cite{shao2024visual} and the agentic cropping in DeepEyes~\cite{zheng2025deepeyes}); and
(ii) RoI-resampling, which extracts RoI-aligned visual features directly from the feature map for re-injection (e.g., VGR~\cite{wang2025vgrvisualgroundedreasoning}).

We then analyze ROVER's components via progressive variants:
(i) $\mathrm{ROVER}\text{-}\mathrm{Sift}_s$ (using standard attention);
(ii) $\mathrm{ROVER}\text{-}\mathrm{Sift}_d$ (using differential attention);
(iii) $\mathrm{ROVER}\text{-}\mathrm{SW}$ (adding \textbf{Weave} and VWS); and
(iv) the full $\mathrm{ROVER}\text{-}\mathrm{LSW}$ (\textbf{Link}+\textbf{Sift}+\textbf{Weave}).
Finally, on MM-GCoT, we report $\mathrm{ROVER}\text{-}\mathrm{LSW}\text{-}\mathrm{GRPO}$ to isolate the gains from reinforcement learning.

\subsection{Multi-Image Reasoning on VideoEspresso}
\label{sec:exp_videoespresso}

\paragraph{Main Results.}
\label{sec:exp_videoespresso_main}
\definecolor{ablagray}{HTML}{F7F7F7}
\definecolor{groupgray}{HTML}{F2F2F2}

\begin{table*}[t]
\centering
\caption{\textbf{Main results on VideoEspresso.}
We report the official VideoEspresso baseline~\cite{Han_2025_CVPR} and our $\mathrm{ROVER}\text{-}\mathrm{LSW}$ in a controlled setting.
$^{\ast}$: full video inputs (otherwise core frames); $^{\dagger}$: LLM-as-a-judge verification after similarity retrieval. \best{Best} and \second{second-best} results are highlighted.
}
\label{tab:videoespresso_main}

\setlength{\tabcolsep}{2.5pt}
\renewcommand{\arraystretch}{1.05}

\resizebox{\textwidth}{!}{%
\begin{tabular}{l|c|c|c|c|c|c|c|c|c|c|c|c|c|c|c|c}
\toprule
\textbf{Method}
& \textbf{Params}
& \textbf{Narra.} & \textbf{Event} & \textbf{Ingre.} & \textbf{Causal} & \textbf{Theme}
& \textbf{Conte.} & \textbf{Influ.} & \textbf{Role} & \textbf{Inter.} & \textbf{Behav.}
& \textbf{Emoti.} & \textbf{Cook.} & \textbf{Traff.} & \textbf{Situa.}
& \textbf{Avg.} \\
\midrule

\rowcolor{groupgray}
\multicolumn{17}{c}{\textit{Closed-source MLLMs}$^{\star}$} \\
\midrule
GPT-4o~\cite{hurst2024gpt}
& --
& 32.3 & 16.7 & 25.5 & 22.8 & 32.8
& 27.5 & 37.5 & 28.6 & 24.2 & 19.3
& 30.8 & 30.2 & 20.0 & 22.0
& 26.4 \\
Qwen-VL-Max~\cite{Bai2023QwenVLAF}
& --
& 33.9 & 22.4 & 23.5 & 21.4 & 26.2
& 30.3 & 41.7 & 30.2 & 27.4 & 26.3
& 20.0 & 20.8 & 16.7 & 24.0
& 26.0 \\

\midrule
\rowcolor{groupgray}
\multicolumn{17}{c}{\textit{Opened-source MLLMs}$^{\star}$} \\
\midrule
LLaVA-1.5~\cite{liu2023llava}
& 6.8B
& 32.3 & 21.3 & 19.4 & 17.1 & 26.2
& 20.2 & 36.1 & 33.3 & 21.0 & 21.1
& 20.0 & 35.8 & 16.7 & 18.0
& 24.2 \\
InternVL2~\cite{chen2024internvl}
& 8.1B
& 33.9 & 24.1 & 27.6 & 24.4 & 42.6
& 33.0 & 45.8 & 28.6 & 19.4 & 22.8
& 21.5 & 34.0 & 20.0 & 24.0
& 28.7 \\
LLaVA-N-Inter~\cite{li2024llava_nextinterleave}
& 8.1B
& 24.2 & 23.6 & 26.5 & 19.2 & 31.1
& 32.1 & 31.9 & 17.5 & 24.2 & 21.1
& 26.2 & 30.2 & 13.3 & 20.0
& 24.4 \\
Qwen2-VL~\cite{wang2024qwen2}
& 8.3B
& 27.4 & 23.0 & 24.5 & 23.5 & 29.5
& 31.2 & 47.2 & 31.7 & 22.6 & \second{28.1}
& \best{40.0} & 22.6 & 30.0 & 18.0
& 28.5 \\
LongVA-DPO~\cite{zhang2024longva}
& 7.9B
& 35.5 & 14.9 & 16.3 & 19.0 & 34.4
& 22.0 & 37.5 & 23.8 & 29.0 & 22.8
& 20.0 & 37.7 & 16.7 & 12.0
& 24.4 \\
mPLUG-Owl3~\cite{ye2024mplug}
& 8.1B
& 30.6 & 23.6 & 20.4 & 22.3 & 37.7
& 29.4 & \second{48.6} & 34.9 & 30.6 & 24.6
& 27.7 & 24.5 & 13.3 & 24.0
& 28.0 \\
LLaVA-N-Video~\cite{zhang2024llavanextvideo}
& 7.1B
& 31.2 & 20.2 & 16.2 & 17.6 & 36.5
& 32.7 & 30.6 & 24.5 & 26.4 & 24.5
& 34.7 & 20.8 & 20.3 & 17.0
& 25.2 \\

\midrule
\rowcolor{groupgray}
\multicolumn{17}{c}{\textit{Baseline}} \\
\midrule
VideoEspresso~\cite{Han_2025_CVPR}
& 8.5B
& 45.2 & 27.0 & 33.7 & 26.1 & 39.3
& 36.7 & \best{55.6} & 41.3 & 30.6 & \best{29.8}
& 30.8 & 35.8 & 20.0 & 26.0
& 34.1 \\

\midrule
\rowcolor{groupgray}
\multicolumn{17}{c}{\textit{Main results on Qwen2.5-VL-7B}} \\
\midrule

+ $\mathrm{ROVER}\text{-}{\mathrm{LSW}}$
& 8.5B
& \best{50.6}  & \best{45.9} & \best{42.9} & \best{33.8} & \best{45.9}
& \best{43.1} & 37.5 & \best{42.9} & \best{41.9} & 26.3
& \second{36.9} & \best{49.1} & \best{56.6} & \best{44.0}
& \best{42.7} \\

\midrule
+ $\mathrm{ROVER}\text{-}{\mathrm{LSW}}^{\dagger}$
& 8.5B
& \second{49.4} & \second{45.2} & \second{41.8} & \second{33.6} & \second{44.3}
& \second{41.3} & 36.1 & \second{41.3} & \second{40.3} & 24.6
& 35.4 & \second{47.2} & \second{53.3} & \second{40.0}
& \second{41.0} \\

\bottomrule
\end{tabular}%
}
\end{table*}

\Cref{tab:videoespresso_main} reports the main results on VideoEspresso.
Overall, $\mathrm{ROVER}\text{-}\mathrm{LSW}$ achieves the best average in our controlled setting, outperforming the previous best baseline by +8.6\%.
For evaluation, we adhere to the benchmark's official similarity-based option-matching protocol. Furthermore, we apply an LLM-as-a-judge verification using OpenAI o3~\cite{openai_o3_o4mini_2025} to assess response--option agreement across logical consistency, factuality, accuracy, and conciseness, complementing the benchmark's subjective evaluation~\cite{Han_2025_CVPR}.
This verification typically lowers absolute scores but yields a more reproducible estimate of reasoning correctness.

\paragraph{Ablations and Analysis.}
\definecolor{ablagray}{HTML}{F7F7F7}
\definecolor{groupgray}{HTML}{F2F2F2}

\begin{table*}[t]
\centering
\caption{\textbf{Ablations on VideoEspresso.}
Comparison against RoI-reencoding/resampling methods and component ablation under identical configurations.
}
\label{tab:videoespresso_ablation}

\setlength{\tabcolsep}{2.5pt}
\renewcommand{\arraystretch}{1.05}

\resizebox{\textwidth}{!}{%
\begin{tabular}{l|c|c|c|c|c|c|c|c|c|c|c|c|c|c|c|c|c}
\toprule
\textbf{Method}
& \textbf{$\Delta$Params} & \textbf{Gen. Toks}
& \textbf{Narra.} & \textbf{Event} & \textbf{Ingre.} & \textbf{Causal} & \textbf{Theme}
& \textbf{Conte.} & \textbf{Influ.} & \textbf{Role} & \textbf{Inter.} & \textbf{Behav.}
& \textbf{Emoti.} & \textbf{Cook.} & \textbf{Traff.} & \textbf{Situa.}
& \textbf{Avg.} \\

\midrule
\rowcolor{groupgray}
\multicolumn{18}{c}{\textit{Baseline}} \\
\midrule
VideoEspresso~\cite{Han_2025_CVPR}
& +0.4B & 448.4
& 45.2 & 27.0 & 33.7 & 26.1 & 39.3
& 36.7 & \best{55.6} & 41.3 & 30.6 & \best{29.8}
& 30.8 & 35.8 & 20.0 & 26.0
& 34.1 \\

\midrule
\rowcolor{groupgray}
\multicolumn{18}{c}{\textit{Ablations on Qwen2.5-VL-7B}} \\
\midrule
+ RoI-reencoding~\cite{man2025argus}
& +0.0B & 686.7
& 41.8 & 42.0 & 39.8 & 32.4 & 41.0
& 39.4 & 34.7 & 39.7 & 38.7 & 22.8
& 32.3 & 39.6 & 46.7 & 36.0
& 37.6 \\
+ RoI-resampling~\cite{man2025argus}
& +0.0B & 732.1
& 44.3 & 40.8 & 41.8 & 33.1 & 42.6
& \second{43.1} & 31.9 & \second{42.9} & 30.6 & 24.6
& 35.4 & 39.6 & 43.3 & 40.0
& 38.1 \\
+ $\mathrm{ROVER}\text{-}{\mathrm{Sift}_s}$
& +0.1B & 318.5
& 46.8 & 38.9 & 33.7 & 31.0 & 36.1
& 41.3 & 34.7 & 42.9 & 25.8 & 19.3
& 32.3 & 43.4 & 46.7 & 40.0
& 36.6 \\
+ $\mathrm{ROVER}\text{-}{\mathrm{Sift}_d}$
& +0.1B & 324.8
& 48.1 & 42.0 & 40.8 & 32.4 & 45.9
& 42.2 & \second{40.3} & 39.7 & 33.9 & 26.3
& 33.8 & 45.3 & 50.0 & 42.0
& 40.2 \\
+ $\mathrm{ROVER}\text{-}{\mathrm{SW}}$
& +0.2B & 303.4
& \second{49.4} & \second{45.2} & \best{43.9} & \second{33.8} & \best{47.5}
& 42.2 & 36.1 & 41.3 & \best{43.5} & 24.6
& \second{35.4} & \second{47.2} & \second{53.3} & \second{44.0}
& \second{42.0} \\

+ $\mathrm{ROVER}\text{-}{\mathrm{LSW}}$
& +0.2B & 308.0
& \best{50.6}  & \best{45.9} & \second{42.9} & \best{33.8} & \second{45.9}
& \best{43.1} & 37.5 & \best{42.9} & \second{41.9} & \second{26.3}
& \best{36.9} & \best{49.1} & \best{56.6} & \best{44.0}
& \best{42.7} \\

\bottomrule
\end{tabular}%
}
\end{table*}

\Cref{tab:videoespresso_ablation} dissects the contribution of each component to multi-image reasoning, reporting performance alongside parameter growth ($\Delta$Params) and the average generated token count (evidence + answer) as a proxy for decoding efficiency. Comparing the two \textbf{Sift} variants (both +0.1B params), the standard attention model (36.6\%) trails the RoI-resampling baseline (38.1\%).
In contrast, differential attention yields a distinct improvement to 40.2\%, underscoring the necessity of distractor suppression during evidence routing.
Incorporating history-aware aggregation via \textbf{Weave} and VWS further lifts the average to 42.0\% (+0.2B), with the full \textbf{Link}+\textbf{Sift}+\textbf{Weave} model achieving the best overall performance of 42.7\%.
Notably, $\mathrm{ROVER}\text{-}\mathrm{LSW}$ reduces total output tokens (both generated and inserted) by 57.9\% and 31.3\% compared to RoI-resampling and the text-only baseline, respectively, indicating that our object-centric evidence routing mechanism effectively enhances both reasoning quality and decoding efficiency.

\subsection{Single-Image Reasoning on MM-GCoT}
\label{sec:exp_mmgcot}

\paragraph{Main Results.}
\definecolor{groupgray}{HTML}{F2F2F2}

\begin{table*}[t]
\centering
\caption{\textbf{Main results on MM-GCoT.}
Comparison with GCoT baselines and GRPO effects.
``AF'' and ``GF'' denote answer-first and grounding-first prompting, respectively~\cite{wu2025grounded}.
}
\label{tab:mmgcot_main}
\resizebox{\textwidth}{!}{%
\begin{tabular}{l|c|c|ccc|ccc|ccc|ccc}
\toprule[1.0pt]
\multicolumn{1}{c|}{\multirow{2}{*}{\textbf{Method}}} &
\multicolumn{1}{c|}{\multirow{2}{*}{\textbf{Params}}} &
\multicolumn{1}{c|}{\multirow{2}{*}{\makecell{\textbf{Prompt} \\ \textbf{Setting}}}} & 
\multicolumn{3}{c|}{\textbf{Attribute}} &
\multicolumn{3}{c|}{\textbf{Judgement}} &
\multicolumn{3}{c|}{\textbf{Object}} &
\multicolumn{3}{c}{\textbf{Average}} \\
\multicolumn{1}{c|}{} &
\multicolumn{1}{c|}{} &
\multicolumn{1}{c|}{} &
\textbf{A-Acc.} & \textbf{G-Acc.} & \textbf{Consist.} &
\textbf{A-Acc.} & \textbf{G-Acc.} & \textbf{Consist.} &
\textbf{A-Acc.} & \textbf{G-Acc.} & \textbf{Consist.} &
\textbf{A-Acc.} & \textbf{G-Acc.} & \textbf{Consist.} \\
\midrule

\rowcolor{groupgray}
\multicolumn{15}{c}{\textit{Opened-source MLLMs with prompting}} \\
\midrule

InternVL2.5-8B~\cite{chen2024expanding} & 8.1B & AF
& 61.9 & 51.9 & 42.6  & 89.3 & 36.4 & 31.5  & 48.3 & 43.2 & 39.4  & 66.5 & 43.8 & 37.8 \\
InternVL2.5-8B~\cite{chen2024expanding} & 8.1B & GF
& 54.5 & 43.1 & 40.7  & 76.9 & 31.9 & 26.1  & 37.1 & 46.2 & 31.1  & 56.2 & 40.4 & 32.6 \\

\midrule
LLaVA-7B~\cite{liu2023llava} & 6.8B & AF
& 68.6 & 9.2  & 8.8   & 83.0 & 11.2 & 11.5  & 58.4 & 9.1  & 9.9   & 70.0 & 9.8  & 10.1 \\
LLaVA-7B~\cite{liu2023llava} & 6.8B & GF
& 59.7 & 6.3  & 5.6   & 82.5 & 0.5  & 0.6   & 35.9 & 5.5  & 9.7   & 59.4 & 4.1  & 5.3 \\
\midrule
LLaVA-13B~\cite{liu2023llava} & 13.1B & AF
& 69.7 & 12.0 & 9.6   & 83.5 & 14.6 & 15.4  & 58.7 & 11.6 & 14.4  & 70.6 & 12.7 & 13.1 \\
LLaVA-13B~\cite{liu2023llava} & 13.1B & GF
& 69.7 & 11.5 & 11.7  & 83.3 & 6.1  & 6.6   & 51.9 & 14.6 & 17.2  & 68.3 & 10.8 & 11.8 \\
\midrule
\rowcolor{groupgray}
\multicolumn{15}{c}{\textit{Baseline}} \\
\midrule
LLaVA-7B GCoT~\cite{wu2025grounded} & 6.8B & GCoT
& 72.8 & 66.7 & 56.1  & 88.3 & 61.7 & 56.9  & 62.3 & 61.7 & 61.3  & 74.5 & 63.3 & 58.1 \\
LLaVA-13B GCoT~\cite{wu2025grounded} & 13.1B & GCoT
& 74.7 & 67.8 & 58.0  & 90.3 & 72.8 & 68.0  & \second{64.1} & 60.8 & 59.3  & 76.4 & 67.1 & 61.8 \\

\midrule
\rowcolor{groupgray}
\multicolumn{15}{c}{\textit{Main results on Qwen2.5-VL-7B}} \\
\midrule
+ $\mathrm{ROVER}\text{-}{\mathrm{LSW}}$ & 8.5B & GCoT
& \second{75.6} & \second{78.0} & \second{66.7} & \second{90.8} & \second{90.3} & \second{81.1} & 63.8 & \second{71.4} & \second{64.8} & \second{76.7} & \second{79.9} & \second{70.8} \\
+ $\mathrm{ROVER}\text{-}{\mathrm{LSW}}\text{-}{\mathrm{GRPO}}$ & 8.5B & GCoT
& \best{76.7} & \best{81.0} & \best{68.0} & \best{95.6} & \best{90.3} & \best{85.9} & \best{71.4} & \best{73.9} & \best{68.9} & \best{81.2} & \best{81.7} & \best{74.3} \\
\bottomrule[1.0pt]
\end{tabular}%
}
\end{table*}
Following official protocols~\cite{wu2025grounded}, \Cref{tab:mmgcot_main} shows that $\mathrm{ROVER}\text{-}\mathrm{LSW}$ surpasses the strongest baseline (LLaVA-13B GCoT) by +12.8\% in grounding accuracy and +9.0\% in consistency, indicating tighter evidence--reasoning alignment.
Furthermore, incorporating GRPO yields additional gains, effectively lifting the overall answer accuracy (+4.8\%) while extending the lead in grounding metrics (+14.6\% G-Acc, +12.5\% Consist.) over the baseline.

\paragraph{Ablations and Analysis.}
\begin{table*}[t]
\centering
\caption{\textbf{Ablations on MM-GCoT.}
Comparison against RoI-reencoding/resampling methods and component ablation under identical configurations.
}
\label{tab:mmgcot_ablation}
\resizebox{\textwidth}{!}{%
\begin{tabular}{l|c|c|ccc|ccc|ccc|ccc}
\toprule[1.0pt]
\multicolumn{1}{c|}{\multirow{2}{*}{\textbf{Method}}} &
\multicolumn{1}{c|}{\multirow{2}{*}{\textbf{Params}}} &
\multicolumn{1}{c|}{\multirow{2}{*}{\makecell{\textbf{Prompt} \\ \textbf{Setting}}}} & 
\multicolumn{3}{c|}{\textbf{Attribute}} &
\multicolumn{3}{c|}{\textbf{Judgement}} &
\multicolumn{3}{c|}{\textbf{Object}} &
\multicolumn{3}{c}{\textbf{Average}} \\
\multicolumn{1}{c|}{} &
\multicolumn{1}{c|}{} &
\multicolumn{1}{c|}{} &
\textbf{A-Acc.} & \textbf{G-Acc.} & \textbf{Consist.} &
\textbf{A-Acc.} & \textbf{G-Acc.} & \textbf{Consist.} &
\textbf{A-Acc.} & \textbf{G-Acc.} & \textbf{Consist.} &
\textbf{A-Acc.} & \textbf{G-Acc.} & \textbf{Consist.} \\
\midrule
\rowcolor{groupgray}
\multicolumn{15}{c}{\textit{Baseline}} \\
\midrule
LLaVA-13B GCoT~\cite{wu2025grounded} & 13.1B & GCoT
& 74.7 & 67.8 & 58.0  & 90.3 & 72.8 & 68.0  & \best{64.1} & 60.8 & 59.3  & 76.4 & 67.1 & 61.8 \\
\midrule

\rowcolor{groupgray}
\multicolumn{15}{c}{\textit{Ablations on Qwen2.5-VL-7B}} \\
\midrule
+ RoI-reencoding~\cite{man2025argus} & 8.3B & GCoT
&75.4 & 63.8 & 55.5 & 90.3 & 69.9 & 65.0 & 58.4 & 60.2 & 56.6 & 74.7 & 64.6 & 59.0 \\
+ RoI-resampling~\cite{man2025argus} & 8.3B & GCoT
& \best{76.7} & 64.7 & 56.8 & 90.3 & 70.4 & 65.5 & 57.4 & 60.5 & 56.5 & 74.8 & 65.2 & 59.6 \\
+ $\mathrm{ROVER}\text{-}{\mathrm{Sift}_s}$ & 8.4B & GCoT
& 74.1 & 59.7 & 53.9 & 88.8 & 71.8 & 65.5 & 60.5 & 58.7 & 56.2 & 74.5 & 63.4 & 58.5 \\
+ $\mathrm{ROVER}\text{-}{\mathrm{Sift}_d}$ & 8.4B & GCoT
& 75.4 & 76.3 & 65.3 & 90.7 & 87.9 & 79.5 & 60.8 & 71.1 & 61.3 & 75.6 & 78.4 & 68.7 \\
+ $\mathrm{ROVER}\text{-}{\mathrm{SW}}$ & 8.5B & GCoT
& 75.4 & \second{78.0} & \second{66.4} & \second{90.8} & \second{90.3} & \second{81.1} & 63.2 & \second{71.4} & \second{64.1} & \second{76.5} & \second{79.9} & \second{70.5} \\
+ $\mathrm{ROVER}\text{-}{\mathrm{LSW}}$ & 8.5B & GCoT
& \second{75.6} & \best{78.0} & \best{66.7} & \best{90.8} & \best{90.3} & \best{81.1} & 
\second{63.8} & \best{71.4} & \best{64.8} & \best{76.7} & \best{79.9} & \best{70.8} \\
\bottomrule[1.0pt]
\end{tabular}%
}
\vspace{-2mm}
\end{table*}

\Cref{tab:mmgcot_ablation} breaks down the impact of each component on MM-GCoT.
Relative to the RoI-resampling baseline, introducing \textbf{Sift} with differential attention improves overall G-Acc by +13.2\%, and Consist. by +9.1\%.
Finally, integrating \textbf{Weave}/VWS and \textbf{Link} contributes to sustained improvements, culminating in the best overall performance across all metrics.

\subsection{Backbone Compatibility and Transferability}
\label{sec:exp_transfer}

\begin{figure}[t]
\centering
\includegraphics[width=\linewidth]{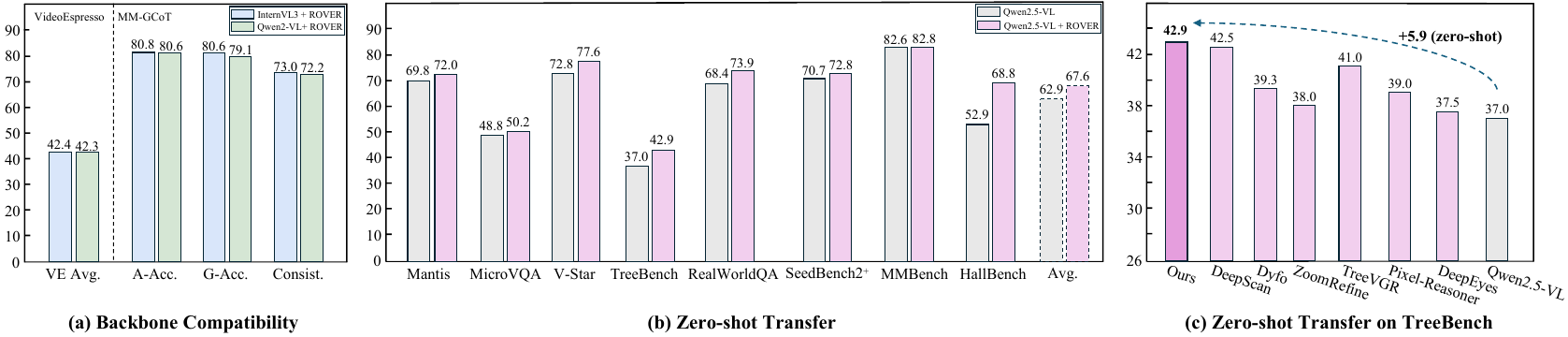}
\caption{\textbf{Backbone compatibility and transferability.}
(a): Backbone compatibility. VideoEspresso (Avg.) evaluated via similarity matching and MM-GCoT (A-Acc./G-Acc./Consist.).
(b): Zero-shot transfer of ROVER-enhanced Qwen2.5-VL-7B to held-out benchmarks.
(c): Comparison with state-of-the-art alternatives on TreeBench~\cite{wang2025traceable}, with scores taken from DeepScan~\cite{li2026deepscan}.
}
\label{fig:transfer}
\end{figure}

As summarized in \Cref{fig:transfer}, we evaluate the versatility of ROVER along two dimensions: 
(i) backbone compatibility, where we extend ROVER to diverse architectures including InternVL3-8B~\cite{zhu2025internvl3} and Qwen2-VL-7B~\cite{wang2024qwen2}; 
and (ii) zero-shot transfer, applying the ROVER-enhanced Qwen2.5-VL-7B (trained on VideoEspresso) to a range of public benchmarks, evaluating generalization to diverse multimodal domains held out during training.

Detailed analysis reveals that ROVER's explicit object-centric routing effectively mitigates hallucinations and enhances spatial understanding,
yielding significant gains on RealWorldQA~\cite{xai2024realworldqa} (+5.5\%) and HallBench~\cite{Guan_2024_CVPR} (+15.9\%).
Moreover, sustained improvements on Mantis~\cite{Jiang2024MANTISIM} (+2.2\%) and MicroVQA~\cite{burgess2025microvqa} (+1.4\%) confirm robust transferability across multi-image reasoning tasks. 
Finally, gains on V-Star~\cite{wu2024v} (+4.8\%), SEED-Bench-2-Plus~\cite{li2024seedbench2plus} (+2.1\%), and TreeBench~\cite{wang2025traceable} (+5.9\%, achieving the best zero-shot result among all compared methods~\cite{li2026deepscan, li2025dyfo, yu2025zoomrefine, wang2025traceable, wang2025pixel, zheng2025deepeyes}) demonstrate that intra-image routing successfully preserves fine-grained visual details and spatial hierarchies to facilitate robust higher-order reasoning.
Collectively, these consistent results confirm that ROVER acquires a highly generalized strategy for object-centric visual evidence routing across diverse domains (see Appendix~\ref{sec:appendix_infercase} for additional qualitative examples).

\section{Conclusion and Discussion}
\label{sec:conclusion}

We presented \textbf{ROVER}, a lightweight learnable plugin that equips autoregressive multimodal generation with a global object-centric pathway for visual evidence routing. Operating via a constant-length \textbf{Link}/\textbf{Sift}/\textbf{Weave} token triplet, ROVER anchors localized objects, extracts complementary context via differential attention while suppressing distractors, and consolidates history-aware evidence across images through a shared Visual Working Space. Trained via an interleaved SFT-to-GRPO pipeline, ROVER achieves consistent gains and establishes new state-of-the-art results on both single-image and multi-image grounded reasoning benchmarks.

\textbf{Discussion 1: Implicit structured representation.} Compressing visual regions into constant-length representations with differential attention potentially acts as an information bottleneck that filters noise and captures underlying scene structure (e.g., semantic correlations and spatial layouts). Conceptually, ROVER mirrors dynamic graph reasoning, where objects act as nodes and the VWS enables implicit message passing, though formal alignment requires future validation.

\textbf{Discussion 2: Adaptive token allocation strategy.} While the constant-length triplet balances efficiency and accuracy, it might constrain information capacity for regions with extreme visual complexity (e.g., dense text). Exploring adaptive routing token allocation conditioned on semantic density could further enhance fine-grained perception. This introduces inevitable trade-offs among representational capacity, latency, and background noise, warranting future investigation.

\textbf{Discussion 3: Parameter footprint optimization.} ROVER guarantees a constant three-token decoding overhead per grounded object with approximately 0.2B additional parameters. Although this footprint is marginal relative to the 7B backbone, exploring even more lightweight architectural designs remains a promising direction for future work.

\clearpage

\bibliographystyle{plain}
\bibliography{reference}

@string {CVPR = "Conference on Computer Vision and Pattern Recognition (CVPR)"}

@string {ACL = "Annual Meeting of the Association for Computational Linguistics (ACL)"}

@string {CHI = "ACM Conference on Human Factors in Computing Systems (CHI)"}

@article{wei2022chain,
  title={Chain-of-thought prompting elicits reasoning in large language models},
  author={Wei, Jason and Wang, Xuezhi and Schuurmans, Dale and Bosma, Maarten and Xia, Fei and Chi, Ed and Le, Quoc V and Zhou, Denny and others},
  journal={Advances in neural information processing systems},
  volume={35},
  pages={24824--24837},
  year={2022}
}

@article{zhang2023multimodal,
  title={Multimodal chain-of-thought reasoning in language models},
  author={Zhang, Zhuosheng and Zhang, Aston and Li, Mu and Zhao, Hai and Karypis, George and Smola, Alex},
  journal={arXiv preprint arXiv:2302.00923},
  year={2023}
}

@article{shao2024visual,
  title={Visual cot: Advancing multi-modal language models with a comprehensive dataset and benchmark for chain-of-thought reasoning},
  author={Shao, Hao and Qian, Shengju and Xiao, Han and Song, Guanglu and Zong, Zhuofan and Wang, Letian and Liu, Yu and Li, Hongsheng},
  journal={Advances in Neural Information Processing Systems},
  volume={37},
  pages={8612--8642},
  year={2024}
}

@article{hu2024visual,
  title={Visual sketchpad: Sketching as a visual chain of thought for multimodal language models},
  author={Hu, Yushi and Shi, Weijia and Fu, Xingyu and Roth, Dan and Ostendorf, Mari and Zettlemoyer, Luke and Smith, Noah A and Krishna, Ranjay},
  journal={arXiv preprint arXiv:2406.09403},
  year={2024}
}

@article{yu2025introducing,
  title={Introducing Visual Perception Token into Multimodal Large Language Model},
  author={Yu, Runpeng and Ma, Xinyin and Wang, Xinchao},
  journal={arXiv preprint arXiv:2502.17425},
  year={2025}
}

@inproceedings{gao2025interleaved,
  title={Interleaved-Modal Chain-of-Thought},
  author={Gao, Jun and Li, Yongqi and Cao, Ziqiang and Li, Wenjie},
  booktitle={2025 IEEE/CVF Conference on Computer Vision and Pattern Recognition (CVPR)},
  pages={19520--19529},
  year={2025},
  organization={IEEE}
}

@article{hurst2024gpt,
  title={Gpt-4o system card},
  author={Hurst, Aaron and Lerer, Adam and Goucher, Adam P and Perelman, Adam and Ramesh, Aditya and Clark, Aidan and Ostrow, AJ and Welihinda, Akila and Hayes, Alan and Radford, Alec and others},
  journal={arXiv preprint arXiv:2410.21276},
  year={2024}
}

@article{shao2024deepseekmath,
  title={Deepseekmath: Pushing the limits of mathematical reasoning in open language models},
  author={Shao, Zhihong and Wang, Peiyi and Zhu, Qihao and Xu, Runxin and Song, Junxiao and Bi, Xiao and Zhang, Haowei and Zhang, Mingchuan and Li, YK and Wu, Yang and others},
  journal={arXiv preprint arXiv:2402.03300},
  year={2024}
}

@article{liu2023llava,
  title={Visual instruction tuning},
  author={Liu, Haotian and Li, Chunyuan and Wu, Qingyang and Lee, Yong Jae},
  journal={Advances in neural information processing systems},
  volume={36},
  pages={34892--34916},
  year={2023}
}

@article{Bai2023QwenVLAF,
  title={Qwen-VL: A Frontier Large Vision-Language Model with Versatile Abilities},
  author={Jinze Bai and Shuai Bai and Shusheng Yang and Shijie Wang and Sinan Tan and Peng Wang and Junyang Lin and Chang Zhou and Jingren Zhou},
  journal={ArXiv},
  year={2023},
  volume={abs/2308.12966},
}

@article{wang2024qwen2,
  title={Qwen2-vl: Enhancing vision-language model's perception of the world at any resolution},
  author={Wang, Peng and Bai, Shuai and Tan, Sinan and Wang, Shijie and Fan, Zhihao and Bai, Jinze and Chen, Keqin and Liu, Xuejing and Wang, Jialin and Ge, Wenbin and others},
  journal={arXiv preprint arXiv:2409.12191},
  year={2024}
}

@misc{chen2025r1v,
  author       = {Chen, Liang and Li, Lei and Zhao, Haozhe and Song, Yifan and Vinci},
  title        = {R1-V: Reinforcing Super Generalization Ability in Vision-Language Models with Less Than \$3},
  howpublished = {\url{https://github.com/Deep-Agent/R1-V}},
  note         = {Accessed: 2025-02-02},
  year         = {2025}
}

@inproceedings{AnthropicModelCA,
  title={Model Card Addendum: Claude 3.5 Haiku and Upgraded Claude 3.5 Sonnet},
  author={Sonnet Anthropic},
  year={2024},
  booktitle={Claude 3.5 Sonnet},
  url={https://api.semanticscholar.org/CorpusID:273639283}
}

@inproceedings{chen2024internvl,
    title={Internvl: Scaling up vision foundation models and aligning for generic visual-linguistic tasks},
    author={Chen, Zhe and Wu, Jiannan and Wang, Wenhai and Su, Weijie and Chen, Guo and Xing, Sen and Zhong, Muyan and Zhang, Qinglong and Zhu, Xizhou and Lu, Lewei and others},
    booktitle={Proceedings of the IEEE/CVF Conference on Computer Vision and Pattern Recognition},
    pages={24185--24198},
    year={2024}
  }

@misc{bai2025qwen25vltechnicalreport,
      title={Qwen2.5-VL Technical Report}, 
      author={Shuai Bai and Keqin Chen and Xuejing Liu and Jialin Wang and Wenbin Ge and Sibo Song and Kai Dang and Peng Wang and Shijie Wang and Jun Tang and Humen Zhong and Yuanzhi Zhu and Mingkun Yang and Zhaohai Li and Jianqiang Wan and Pengfei Wang and Wei Ding and Zheren Fu and Yiheng Xu and Jiabo Ye and Xi Zhang and Tianbao Xie and Zesen Cheng and Hang Zhang and Zhibo Yang and Haiyang Xu and Junyang Lin},
      year={2025},
      eprint={2502.13923},
      archivePrefix={arXiv},
      primaryClass={cs.CV},
      url={https://arxiv.org/abs/2502.13923}, 
}

@article{kojima2022large,
  title={Large language models are zero-shot reasoners},
  author={Kojima, Takeshi and Gu, Shixiang Shane and Reid, Machel and Matsuo, Yutaka and Iwasawa, Yusuke},
  journal={Advances in neural information processing systems},
  volume={35},
  pages={22199--22213},
  year={2022}
}

@article{team2023gemini,
  title={Gemini: a family of highly capable multimodal models},
  author={Gemini Team, Google},
  journal={arXiv preprint arXiv:2312.11805},
  year={2023}
}

@article{li2024llava-ov,
  title={LLaVA-OneVision: Easy Visual Task Transfer},
  author={Li, Bo and Zhang, Yuanhan and Guo, Dong and Zhang, Renrui and Li, Feng and Zhang, Hao and Zhang, Kaichen and Li, Yanwei and Liu, Ziwei and Li, Chunyuan},
  journal={arXiv preprint arXiv:2408.03326},
  year={2024}
}

@misc{openai_o3_o4mini_2025,
  author       = {OpenAI},
  title        = {OpenAI o3},
  year         = {2025},
  howpublished = {\url{https://openai.com/index/introducing-o3-and-o4-mini}},
}

@article{lu2023mathvista,
  title={Mathvista: Evaluating mathematical reasoning of foundation models in visual contexts},
  author={Lu, Pan and Bansal, Hritik and Xia, Tony and Liu, Jiacheng and Li, Chunyuan and Hajishirzi, Hannaneh and Cheng, Hao and Chang, Kai-Wei and Galley, Michel and Gao, Jianfeng},
  journal={arXiv preprint arXiv:2310.02255},
  year={2023}
}

@article{shi2025eagle,
  title={Eagle: Exploring the design space for multimodal llms with mixture of encoders},
  author={Shi, Min and Liu, Fuxiao and Wang, Shihao and Liao, Shijia and Radhakrishnan, Subhashree and Zhao, Yilin and Huang, De-An and Yin, Hongxu and Sapra, Karan and Yacoob, Yaser and others},
  journal={arXiv preprint arXiv:2408.15998},
  year={2024}
}

@inproceedings{he2025mmboundaryadvancingmllmknowledge,
    title = "{MMB}oundary: Advancing {MLLM} Knowledge Boundary Awareness through Reasoning Step Confidence Calibration",
    author = "He, Zhitao  and
      Polisetty, Sandeep  and
      Fan, Zhiyuan  and
      Huang, Yuchen  and
      Wu, Shujin  and
      Fung, Yi R.",
    booktitle = "Proceedings of the 63rd Annual Meeting of the Association for Computational Linguistics (Volume 1: Long Papers)",
    month = jul,
    year = "2025",
    address = "Vienna, Austria",
    publisher = "Association for Computational Linguistics",
    url = "https://aclanthology.org/2025.acl-long.802/",
    doi = "10.18653/v1/2025.acl-long.802",
    pages = "16427--16444",
    ISBN = "979-8-89176-251-0",
}

@article{shen2025satori,
  title={Satori-r1: Incentivizing multimodal reasoning with spatial grounding and verifiable rewards},
  author={Shen, Chuming and Wei, Wei and Qu, Xiaoye and Cheng, Yu},
  journal={arXiv preprint arXiv:2505.19094},
  year={2025}
}

@article{tong2024cambrian,
  title={Cambrian-1: A fully open, vision-centric exploration of multimodal llms},
  author={Tong, Peter and Brown, Ellis and Wu, Penghao and Woo, Sanghyun and IYER, Adithya Jairam Vedagiri and Akula, Sai Charitha and Yang, Shusheng and Yang, Jihan and Middepogu, Manoj and Wang, Ziteng and others},
  journal={Advances in Neural Information Processing Systems},
  volume={37},
  pages={87310--87356},
  year={2024}
}

@inproceedings{tong2024eyes,
  title={Eyes wide shut? exploring the visual shortcomings of multimodal llms},
  author={Tong, Shengbang and Liu, Zhuang and Zhai, Yuexiang and Ma, Yi and LeCun, Yann and Xie, Saining},
  booktitle={Proceedings of the IEEE/CVF Conference on Computer Vision and Pattern Recognition},
  pages={9568--9578},
  year={2024}
}

@article{wang2025simpleo3,
  title={Simple o3: Towards interleaved vision-language reasoning},
  author={Wang, Ye and Chen, Qianglong and Li, Zejun and Wang, Siyuan and Guo, Shijie and Zhang, Zhirui and Wei, Zhongyu},
  journal={arXiv preprint arXiv:2508.12109},
  year={2025}
}

@article{wang2025vgrvisualgroundedreasoning,
  title={Vgr: Visual grounded reasoning},
  author={Wang, Jiacong and Kang, Zijian and Wang, Haochen and Jiang, Haiyong and Li, Jiawen and Wu, Bohong and Wang, Ya and Ran, Jiao and Liang, Xiao and Feng, Chao and others},
  journal={arXiv preprint arXiv:2506.11991},
  year={2025}
}

@inproceedings{man2025argus,
  title={Argus: Vision-Centric Reasoning with Grounded Chain-of-Thought},
  author={Man, Yunze and Huang, De-An and Liu, Guilin and Sheng, Shiwei and Liu, Shilong and Gui, Liang-Yan and Kautz, Jan and Wang, Yu-Xiong and Yu, Zhiding},
  booktitle={Proceedings of the Computer Vision and Pattern Recognition Conference},
  pages={14268--14280},
  year={2025}
}

@article{chen2025mint,
  title={MINT-CoT: Enabling Interleaved Visual Tokens in Mathematical Chain-of-Thought Reasoning},
  author={Chen, Xinyan and Zhang, Renrui and Jiang, Dongzhi and Zhou, Aojun and Yan, Shilin and Lin, Weifeng and Li, Hongsheng},
  journal={arXiv preprint arXiv:2506.05331},
  year={2025}
}

@article{oberauer2019working,
  title={Working memory and attention--A conceptual analysis and review},
  author={Oberauer, Klaus},
  journal={Journal of cognition},
  volume={2},
  number={1},
  pages={36},
  year={2019}
}

@article{zhang2022look,
  title={Look twice: A generalist computational model predicts return fixations across tasks and species},
  author={Zhang, Mengmi and Armendariz, Marcelo and Xiao, Will and Rose, Olivia and Bendtz, Katarina and Livingstone, Margaret and Ponce, Carlos and Kreiman, Gabriel},
  journal={PLoS computational biology},
  volume={18},
  number={11},
  pages={e1010654},
  year={2022},
  publisher={Public Library of Science San Francisco, CA USA}
}

@article{lewis2018removal,
  title={The removal of information from working memory},
  author={Lewis-Peacock, Jarrod A and Kessler, Yoav and Oberauer, Klaus},
  journal={Annals of the New York Academy of Sciences},
  volume={1424},
  number={1},
  pages={33--44},
  year={2018},
  publisher={Wiley Online Library}
}

@article{nikolaev2025refixation,
  title={Refixation behavior in naturalistic viewing: Methods, mechanisms, and neural correlates},
  author={Nikolaev, Andrey R and Meghanathan, Radha Nila and van Leeuwen, Cees},
  journal={Attention, Perception, \& Psychophysics},
  volume={87},
  number={1},
  pages={25--49},
  year={2025},
  publisher={Springer}
}

@article{wu2025grounded,
  title={Grounded chain-of-thought for multimodal large language models},
  author={Wu, Qiong and Yang, Xiangcong and Zhou, Yiyi and Fang, Chenxin and Song, Baiyang and Sun, Xiaoshuai and Ji, Rongrong},
  journal={arXiv preprint arXiv:2503.12799},
  year={2025}
}

@InProceedings{Han_2025_CVPR,
    author    = {Han, Songhao and Huang, Wei and Shi, Hairong and Zhuo, Le and Su, Xiu and Zhang, Shifeng and Zhou, Xu and Qi, Xiaojuan and Liao, Yue and Liu, Si},
    title     = {VideoEspresso: A Large-Scale Chain-of-Thought Dataset for Fine-Grained Video Reasoning via Core Frame Selection},
    booktitle = {Proceedings of the Computer Vision and Pattern Recognition Conference (CVPR)},
    month     = {June},
    year      = {2025},
    pages     = {26181-26191}
}

@article{Jiang2024MANTISIM,
  title={Mantis: Interleaved multi-image instruction tuning},
  author={Jiang, Dongfu and He, Xuan and Zeng, Huaye and Wei, Cong and Ku, Max and Liu, Qian and Chen, Wenhu},
  journal={arXiv preprint arXiv:2405.01483},
  year={2024}
}

@inproceedings{wu2024v,
  title={V*: Guided Visual Search as a Core Mechanism in Multimodal LLMs},
  author={Wu, Penghao and Xie, Saining},
  booktitle={2024 IEEE/CVF Conference on Computer Vision and Pattern Recognition (CVPR)},
  pages={13084--13094},
  year={2024},
  organization={IEEE}
}

@InProceedings{Guan_2024_CVPR,
    author    = {Guan, Tianrui and Liu, Fuxiao and Wu, Xiyang and Xian, Ruiqi and Li, Zongxia and Liu, Xiaoyu and Wang, Xijun and Chen, Lichang and Huang, Furong and Yacoob, Yaser and Manocha, Dinesh and Zhou, Tianyi},
    title     = {HallusionBench: An Advanced Diagnostic Suite for Entangled Language Hallucination and Visual Illusion in Large Vision-Language Models},
    booktitle = {Proceedings of the IEEE/CVF Conference on Computer Vision and Pattern Recognition (CVPR)},
    month     = {June},
    year      = {2024},
    pages     = {14375-14385}
}

@article{zhang2022automatic,
  title={Automatic chain of thought prompting in large language models},
  author={Zhang, Zhuosheng and Zhang, Aston and Li, Mu and Smola, Alex},
  journal={arXiv preprint arXiv:2210.03493},
  year={2022}
}

@article{schulman2017proximal,
  title={Proximal policy optimization algorithms},
  author={Schulman, John and Wolski, Filip and Dhariwal, Prafulla and Radford, Alec and Klimov, Oleg},
  journal={arXiv preprint arXiv:1707.06347},
  year={2017}
}

@inproceedings{rein2024gpqa,
  title={Gpqa: A graduate-level google-proof q\&a benchmark},
  author={Rein, David and Hou, Betty Li and Stickland, Asa Cooper and Petty, Jackson and Pang, Richard Yuanzhe and Dirani, Julien and Michael, Julian and Bowman, Samuel R},
  booktitle={First Conference on Language Modeling},
  year={2024}
}

@inproceedings{lightman2023let,
  title={Let's verify step by step},
  author={Lightman, Hunter and Kosaraju, Vineet and Burda, Yuri and Edwards, Harrison and Baker, Bowen and Lee, Teddy and Leike, Jan and Schulman, John and Sutskever, Ilya and Cobbe, Karl},
  booktitle={The Twelfth International Conference on Learning Representations},
  year={2023}
}

@article{jain2024livecodebench,
  title={Livecodebench: Holistic and contamination free evaluation of large language models for code},
  author={Jain, Naman and Han, King and Gu, Alex and Li, Wen-Ding and Yan, Fanjia and Zhang, Tianjun and Wang, Sida and Solar-Lezama, Armando and Sen, Koushik and Stoica, Ion},
  journal={arXiv preprint arXiv:2403.07974},
  year={2024}
}

@inproceedings{fu2024blink,
  title={Blink: Multimodal large language models can see but not perceive},
  author={Fu, Xingyu and Hu, Yushi and Li, Bangzheng and Feng, Yu and Wang, Haoyu and Lin, Xudong and Roth, Dan and Smith, Noah A and Ma, Wei-Chiu and Krishna, Ranjay},
  booktitle={European Conference on Computer Vision},
  pages={148--166},
  year={2024},
  organization={Springer}
}

@article{achiam2023gpt,
  title={Gpt-4 technical report},
  author={Achiam, Josh and Adler, Steven and Agarwal, Sandhini and Ahmad, Lama and Akkaya, Ilge and Aleman, Florencia Leoni and Almeida, Diogo and Altenschmidt, Janko and Altman, Sam and Anadkat, Shyamal and others},
  journal={arXiv preprint arXiv:2303.08774},
  year={2023}
}

@article{ye2024differential,
  title={Differential transformer},
  author={Ye, Tianzhu and Dong, Li and Xia, Yuqing and Sun, Yutao and Zhu, Yi and Huang, Gao and Wei, Furu},
  journal={arXiv preprint arXiv:2410.05258},
  year={2024}
}

@article{yang2023mm,
  title={Mm-react: Prompting chatgpt for multimodal reasoning and action},
  author={Yang, Zhengyuan and Li, Linjie and Wang, Jianfeng and Lin, Kevin and Azarnasab, Ehsan and Ahmed, Faisal and Liu, Zicheng and Liu, Ce and Zeng, Michael and Wang, Lijuan},
  journal={arXiv preprint arXiv:2303.11381},
  year={2023}
}

@inproceedings{zhao2024chatspot,
  title={ChatSpot: bootstrapping multimodal LLMs via precise referring instruction tuning},
  author={Zhao, Liang and Yu, En and Ge, Zheng and Yang, Jinrong and Wei, Haoran and Zhou, Hongyu and Sun, Jianjian and Peng, Yuang and Dong, Runpei and Han, Chunrui and others},
  booktitle={Proceedings of the Thirty-Third International Joint Conference on Artificial Intelligence},
  pages={1743--1752},
  year={2024}
}

@article{su2025thinking,
  title={Thinking with images for multimodal reasoning: Foundations, methods, and future frontiers},
  author={Su, Zhaochen and Xia, Peng and Guo, Hangyu and Liu, Zhenhua and Ma, Yan and Qu, Xiaoye and Liu, Jiaqi and Li, Yanshu and Zeng, Kaide and Yang, Zhengyuan and others},
  journal={arXiv preprint arXiv:2506.23918},
  year={2025}
}

@article{zheng2025deepeyes,
  title={Deepeyes: Incentivizing" thinking with images" via reinforcement learning},
  author={Zheng, Ziwei and Yang, Michael and Hong, Jack and Zhao, Chenxiao and Xu, Guohai and Yang, Le and Shen, Chao and Yu, Xing},
  journal={arXiv preprint arXiv:2505.14362},
  year={2025}
}

@inproceedings{gupta2023visual,
  title={Visual programming: Compositional visual reasoning without training},
  author={Gupta, Tanmay and Kembhavi, Aniruddha},
  booktitle={Proceedings of the IEEE/CVF conference on computer vision and pattern recognition},
  pages={14953--14962},
  year={2023}
}

@article{li2024llava_nextinterleave,
  title={Llava-next-interleave: Tackling multi-image, video, and 3d in large multimodal models},
  author={Li, Feng and Zhang, Renrui and Zhang, Hao and Zhang, Yuanhan and Li, Bo and Li, Wei and Ma, Zejun and Li, Chunyuan},
  journal={arXiv preprint arXiv:2407.07895},
  year={2024}
}

@article{zhang2024longva,
  title={Long context transfer from language to vision},
  author={Zhang, Peiyuan and Zhang, Kaichen and Li, Bo and Zeng, Guangtao and Yang, Jingkang and Zhang, Yuanhan and Wang, Ziyue and Tan, Haoran and Li, Chunyuan and Liu, Ziwei},
  journal={arXiv preprint arXiv:2406.16852},
  year={2024}
}

@article{ye2024mplug,
  title={mplug-owl3: Towards long image-sequence understanding in multi-modal large language models},
  author={Ye, Jiabo and Xu, Haiyang and Liu, Haowei and Hu, Anwen and Yan, Ming and Qian, Qi and Zhang, Ji and Huang, Fei and Zhou, Jingren},
  journal={arXiv preprint arXiv:2408.04840},
  year={2024}
}

@misc{zhang2024llavanextvideo,
	title        = {LLaVA-NeXT: A Strong Zero-shot Video Understanding Model},
	author       = {Zhang, Yuanhan and Li, Bo and Liu, haotian and Lee, Yong jae and Gui, Liangke and Fu, Di and Feng, Jiashi and Liu, Ziwei and Li, Chunyuan},
	year         = 2024,
	month        = {April},
	url          = {https://llava-vl.github.io/blog/2024-04-30-llava-next-video/}
}

@inproceedings{bge-m3,
    title = "{M}3-Embedding: Multi-Linguality, Multi-Functionality, Multi-Granularity Text Embeddings Through Self-Knowledge Distillation",
    author = "Chen, Jianlyu  and
      Xiao, Shitao  and
      Zhang, Peitian  and
      Luo, Kun  and
      Lian, Defu  and
      Liu, Zheng",
    booktitle = "Findings of the Association for Computational Linguistics: ACL 2024",
    month = aug,
    year = "2024",
    address = "Bangkok, Thailand",
    publisher = "Association for Computational Linguistics",
    url = "https://aclanthology.org/2024.findings-acl.137/",
    doi = "10.18653/v1/2024.findings-acl.137",
    pages = "2318--2335"
}

@article{vaswani2017attention,
  title={Attention is all you need},
  author={Vaswani, Ashish and Shazeer, Noam and Parmar, Niki and Uszkoreit, Jakob and Jones, Llion and Gomez, Aidan N and Kaiser, {\L}ukasz and Polosukhin, Illia},
  journal={Advances in neural information processing systems},
  volume={30},
  year={2017}
}

@inproceedings{zheng2024llamafactory,
  title={LlamaFactory: Unified Efficient Fine-Tuning of 100+ Language Models},
  author={Yaowei Zheng and Richong Zhang and Junhao Zhang and Yanhan Ye and Zheyan Luo and Zhangchi Feng and Yongqiang Ma},
  booktitle={Proceedings of the 62nd Annual Meeting of the Association for Computational Linguistics (Volume 3: System Demonstrations)},
  address={Bangkok, Thailand},
  publisher={Association for Computational Linguistics},
  year={2024},
  url={http://arxiv.org/abs/2403.13372}
}

@inproceedings{burgess2025microvqa,
  title={Microvqa: A multimodal reasoning benchmark for microscopy-based scientific research},
  author={Burgess, James and Nirschl, Jeffrey J and Bravo-S{\'a}nchez, Laura and Lozano, Alejandro and Gupte, Sanket Rajan and Galaz-Montoya, Jesus G and Zhang, Yuhui and Su, Yuchang and Bhowmik, Disha and Coman, Zachary and others},
  booktitle={Proceedings of the IEEE/CVF Conference on Computer Vision and Pattern Recognition},
  pages={19552--19564},
  year={2025}
}

@article{zhu2025internvl3,
  title={Internvl3: Exploring advanced training and test-time recipes for open-source multimodal models},
  author={Zhu, Jinguo and Wang, Weiyun and Chen, Zhe and Liu, Zhaoyang and Ye, Shenglong and Gu, Lixin and Tian, Hao and Duan, Yuchen and Su, Weijie and Shao, Jie and others},
  journal={arXiv preprint arXiv:2504.10479},
  year={2025}
}

@misc{xai2024realworldqa,
  title={RealWorldQA: A Benchmark for Evaluating Spatial Understanding and Physical Reasoning in the Real World},
  author={xAI},
  year={2024},
  url={https://x.ai/blog/grok-1.5v},
  note={Benchmark release}
}

@article{zhong2025focus,
  title={FOCUS: Internal MLLM representations for efficient fine-grained visual question answering},
  author={Zhong, Liangyu and Rosenthal, Fabio and Sicking, Joachim and H{\"u}ger, Fabian and Bagdonat, Thorsten and Gottschalk, Hanno and Schwinn, Leo},
  journal={arXiv preprint arXiv:2506.21710},
  year={2025}
}

@article{li2026deepscan,
  title={DeepScan: A Training-Free Framework for Visually Grounded Reasoning in Large Vision-Language Models},
  author={Li, Yangfu and Zhan, Hongjian and Chen, Jiawei and Gong, Yuning and Liu, Qi and Lu, Yue},
  journal={arXiv preprint arXiv:2603.03857},
  year={2026}
}

@article{qiao2025v,
  title={V-thinker: Interactive thinking with images},
  author={Qiao, Runqi and Tan, Qiuna and Yang, Minghan and Dong, Guanting and Yang, Peiqing and Lang, Shiqiang and Wan, Enhui and Wang, Xiaowan and Xu, Yida and Yang, Lan and others},
  journal={arXiv preprint arXiv:2511.04460},
  year={2025}
}

@article{bai2025multi,
  title={Multi-step visual reasoning with visual tokens scaling and verification},
  author={Bai, Tianyi and Hu, Zengjie and Sun, Fupeng and Qiu, Jiantao and Jiang, Yizhen and He, Guangxin and Zeng, Bohan and He, Conghui and Yuan, Binhang and Zhang, Wentao},
  journal={arXiv preprint arXiv:2506.07235},
  year={2025}
}

@article{wang2025pixel,
  title={Pixel reasoner: Incentivizing pixel-space reasoning with curiosity-driven reinforcement learning},
  author={Wang, Haozhe and Su, Alex and Ren, Weiming and Lin, Fangzhen and Chen, Wenhu},
  journal={arXiv preprint arXiv:2505.15966},
  year={2025}
}

@article{wang2025traceable,
  title={Traceable evidence enhanced visual grounded reasoning: Evaluation and methodology},
  author={Wang, Haochen and Li, Xiangtai and Huang, Zilong and Wang, Anran and Wang, Jiacong and Zhang, Tao and Zheng, Jiani and Bai, Sule and Kang, Zijian and Feng, Jiashi and others},
  journal={arXiv preprint arXiv:2507.07999},
  year={2025}
}

@misc{li2024seedbench2plus,
      title={SEED-Bench-2-Plus: Benchmarking Multimodal Large Language Models with Text-Rich Visual Comprehension}, 
      author={Bohao Li and Yuying Ge and Yi Chen and Yixiao Ge and Ruimao Zhang and Ying Shan},
      year={2024},
      eprint={2404.16790},
      archivePrefix={arXiv},
      primaryClass={cs.CV},
      url={https://arxiv.org/abs/2404.16790}, 
}

@inproceedings{li2025dyfo,
  title={Dyfo: A training-free dynamic focus visual search for enhancing lmms in fine-grained visual understanding},
  author={Li, Geng and Xu, Jinglin and Zhao, Yunzhen and Peng, Yuxin},
  booktitle={Proceedings of the Computer Vision and Pattern Recognition Conference},
  pages={9098--9108},
  year={2025}
}

@misc{yu2025zoomrefine,
      title={Zoom-Refine: Boosting High-Resolution Multimodal Understanding via Localized Zoom and Self-Refinement}, 
      author={Xuan Yu and Dayan Guan and Yanfeng Gu},
      year={2025},
      eprint={2506.01663},
      archivePrefix={arXiv},
      primaryClass={cs.CV},
      url={https://arxiv.org/abs/2506.01663}, 
}

@article{kahneman2011thinking,
  title={Thinking, fast and slow},
  author={Kahneman, Daniel},
  journal={Farrar, Straus and Giroux},
  year={2011}
}

@inproceedings{liu2024grounding,
  title={Grounding dino: Marrying dino with grounded pre-training for open-set object detection},
  author={Liu, Shilong and Zeng, Zhaoyang and Ren, Tianhe and Li, Feng and Zhang, Hao and Yang, Jie and Jiang, Qing and Li, Chunyuan and Yang, Jianwei and Su, Hang and others},
  booktitle={European conference on computer vision},
  pages={38--55},
  year={2024},
  organization={Springer}
}

@inproceedings{he2016deep,
  title={Deep residual learning for image recognition},
  author={He, Kaiming and Zhang, Xiangyu and Ren, Shaoqing and Sun, Jian},
  booktitle={Proceedings of the IEEE conference on computer vision and pattern recognition},
  pages={770--778},
  year={2016}
}

@inproceedings{chen2025ict,
  title={Ict: Image-object cross-level trusted intervention for mitigating object hallucination in large vision-language models},
  author={Chen, Junzhe and Zhang, Tianshu and Huang, Shiyu and Niu, Yuwei and Zhang, Linfeng and Wen, Lijie and Hu, Xuming},
  booktitle={Proceedings of the Computer Vision and Pattern Recognition Conference},
  pages={4209--4221},
  year={2025}
}

@inproceedings{guo2024regiongpt,
  title={Regiongpt: Towards region understanding vision language model},
  author={Guo, Qiushan and De Mello, Shalini and Yin, Hongxu and Byeon, Wonmin and Cheung, Ka Chun and Yu, Yizhou and Luo, Ping and Liu, Sifei},
  booktitle={Proceedings of the IEEE/CVF Conference on Computer Vision and Pattern Recognition},
  pages={13796--13806},
  year={2024}
}

@inproceedings{li2024llama,
  title={Llama-vid: An image is worth 2 tokens in large language models},
  author={Li, Yanwei and Wang, Chengyao and Jia, Jiaya},
  booktitle={European Conference on Computer Vision},
  pages={323--340},
  year={2024},
  organization={Springer}
}

@inproceedings{li2025migician,
  title={Migician: Revealing the magic of free-form multi-image grounding in multimodal large language models},
  author={Li, You and Huang, Heyu and Chen, Chi and Huang, Kaiyu and Huang, Chao and Guo, Zonghao and Liu, Zhiyuan and Xu, Jinan and Li, Yuhua and Li, Ruixuan and others},
  booktitle={Findings of the Association for Computational Linguistics: ACL 2025},
  pages={9845--9867},
  year={2025}
}

@article{loshchilov2017adamw,
  title={Decoupled weight decay regularization},
  author={Loshchilov, Ilya and Hutter, Frank},
  journal={arXiv preprint arXiv:1711.05101},
  year={2017}
}

@article{chen2024expanding,
title={Expanding Performance Boundaries of Open-Source Multimodal Models with Model, Data, and Test-Time Scaling},
author={Chen, Zhe and Wang, Weiyun and Cao, Yue and Liu, Yangzhou and Gao, Zhangwei and Cui, Erfei and Zhu, Jinguo and Ye, Shenglong and Tian, Hao and Liu, Zhaoyang and others},
journal={arXiv preprint arXiv:2412.05271},
year={2024}
}
\clearpage
\newpage
\appendix
\section{Appendix}

\subsection{Additional Implementation Details}
\label{sec:appendix_impl}

\paragraph{Training Details.}
All SFT and GRPO experiments are conducted on eight NVIDIA A800 (80GB) GPUs. We implement SFT using LLaMA-Factory~\cite{zheng2024llamafactory} and GRPO using R1-V~\cite{chen2025r1v}. We employ the AdamW optimizer~\cite{loshchilov2017adamw} across all stages.
For interleaved SFT on VideoEspresso, we train for 1 epoch with a cosine scheduler, a peak learning rate of $1\times10^{-5}$, a warmup ratio of 0.1, and an effective global batch size of 64. MM-GCoT SFT uses the same setting except for a higher peak learning rate of $2\times10^{-5}$. Subsequently, for MM-GCoT GRPO, we train on a disjoint 14K split for 2 epochs with a linearly decayed learning rate (starting from $3\times10^{-6}$), an effective global batch size of 64, and $G=4$ sampled generations per prompt. The reward is set to 1 only if the generated output is parsable and the final answer is correct (otherwise 0). The KL coefficient $\beta$ is set to 0.04. 

We formally present our core operational procedures via pseudocode. We detail the object-centric evidence injection in our decoding pipeline (\Cref{alg:rover_decode}) and summarize the trajectory-level optimization in the interleaved GRPO loop (\Cref{alg:rover_grpo}).

\begin{algorithm}[htbp]
\caption{ROVER Decoding Pipeline}
\label{alg:rover_decode}
\begin{algorithmic}[1]
\Require Images $\mathcal{I} = \{\mathcal{I}_m\}_{m=1}^M$, Instruction prompt $\mathbf{X}_{\mathrm{prompt}}$, Vision Encoder, LLM
\Ensure Generated multimodal response $\mathbf{Y}$
\State Extract image tokens $\{\mathbf{V}_m\}$
\State Initialize $\mathcal{W}_0 \gets \varnothing$, $k \gets 0$, $\mathbf{s} \gets \mathbf{X}_{\mathrm{prompt}}$
\While{generation is not finished}
    \State $y_t \gets \text{LLM.GenerateNextToken}(\mathbf{s}, \{\mathbf{V}_m\})$
    \State Append $y_t$ to $\mathbf{s}$
    
    \If{$y_t$ completes a valid grounding pattern $o_k$ with bounding box $\mathbf{b}_k$}
        \State $k \gets k + 1$
        \State $\mathbf{q}_k \gets \mathrm{AvgPool}\big(\mathbf{V}_{m(k)}[\Omega(\mathbf{b}_k)]\big)$ \Comment{Extract RoI summary}
        \State $\mathbf{C}_k \gets \mathbf{V}_{m(k)}[\bar{\Omega}(\mathbf{b}_k)]$ \Comment{Obtain non-RoI context}
        
        \State $\mathbf{t}^{\mathrm{Sift}}_k \gets \mathbf{q}_k$ \textbf{if} $|\mathbf{C}_k| = 0$ \textbf{else} $\mathrm{Enc}_{\mathrm{Sift}}\big(\mathbf{q}_k;\ \mathbf{C}_k,\ \mathbf{M}^{\mathrm{Sift}}_k\big)$ \Comment{DiffAttn with fallback}
        
        \State $\mathcal{W}_k \gets [\mathcal{W}_{k-1};\ \mathbf{t}^{\mathrm{Sift}}_k]$ \Comment{Update VWS memory}
        \State $\mathbf{t}^{\mathrm{Weave}}_k \gets \mathrm{Enc}_{\mathrm{Weave}}\big(\mathbf{t}^{\mathrm{Sift}}_k;\ \mathcal{W}_k,\ \mathbf{M}^{\mathrm{Weave}}_k\big)$ \Comment{Route inter-object history}
        \State $\mathbf{T}_k \gets [\mathbf{t}^{\mathrm{Link}}_k;\ \mathbf{t}^{\mathrm{Sift}}_k;\ \mathbf{t}^{\mathrm{Weave}}_k]$ \Comment{Form constant-length triplet}
        \State Append $\mathbf{T}_k$ to $\mathbf{s}$ \Comment{Inject triplet into sequence}
    \EndIf
\EndWhile
\State \Return $\mathbf{Y} \gets \mathbf{s}$
\end{algorithmic}
\end{algorithm}
\begin{algorithm}[htbp]
\caption{Interleaved GRPO for ROVER}
\label{alg:rover_grpo}
\begin{algorithmic}[1]
\Require Initial policy $p_{\theta}$, Reference policy $p_{\mathrm{ref}}$, Dataset $\mathcal{D}$, Group size $G$, KL coef. $\beta$
\Ensure Optimized policy $p_{\theta}$
\While{training is not converged}
    \State Sample a batch of prompts from $\mathcal{D}$
    \State Synchronize old policy $p_{\theta_{\mathrm{old}}} \gets p_{\theta}$
    \For{each prompt}
        \State Sample $G$ interleaved trajectories $\{\tau_i\}_{i=1}^G \sim p_{\theta_{\mathrm{old}}}$
        
        \State Compute binary rewards $\{r(\tau_i)\}_{i=1}^G$ and normalized advantages $\hat{A}(\tau_i)$
        
        \For{each token $y_t \in \tau_i$}
            \State $m_t \gets 0$ if $y_t \in \{\textbf{Link}, \textbf{Sift}, \textbf{Weave}\}$, else $m_t \gets 1$
        \EndFor
        
        \State Compute masked loss $\mathcal{L}_{\mathrm{GRPO}}$ with $m_t$, $\hat{A}(\tau_i)$, $\beta$, and policies ($p_{\theta}, p_{\theta_{\mathrm{old}}}, p_{\mathrm{ref}}$)
    \EndFor
    \State Update parameters $\theta$ by optimizing $\mathcal{L}_{\mathrm{GRPO}}$
\EndWhile
\State \Return $p_{\theta}$
\end{algorithmic}
\end{algorithm}
\paragraph{VideoEspresso Data Preprocessing.}
To optimize interleaving efficiency, we address instances where a single object mention maps to multiple candidate bounding boxes by retaining only the largest box as the representative anchor. This representative box serves as the grounding pattern trigger (\Cref{sec:ROVER_trigger}), thereby minimizing token redundancy in crowded scenes. Notably, complementary information from unselected boxes can be implicitly retrieved via differential attention.

During SFT, we employ diverse instruction templates~\cite{Jiang2024MANTISIM,li2025migician} to prevent format overfitting and enhance instruction-following adaptability, ensuring the learned routing capabilities transfer robustly to broader open-ended scenarios.
As shown in \Cref{fig:case}, the output is structured into an initial reasoning sequence (integrating visual evidence text, grounding tokens, and ROVER triplets) followed by a pure-text final answer.

\subsection{Additional Evaluation Details}
\label{sec:appendix_eval}
For VideoEspresso, we report two evaluation variants.
(i) LLM-as-a-judge verification: we follow the official similarity-based matching with bge-m3~\cite{bge-m3} to retrieve the closest option, and then verify response--option agreement using OpenAI o3~\cite{openai_o3_o4mini_2025} in terms of logical consistency, factuality, accuracy, and conciseness~\cite{Han_2025_CVPR} (see the full prompt in \Cref{fig:judge_prompt}).
(ii) Multiple-choice evaluation: we additionally report direct multiple-choice QA on the test split (\Cref{tab:videoespresso_mcq}). The ROVER-enhanced model outperforms the Qwen2.5-VL-7B by +4.3\%, indicating our gains are robust to the evaluation protocol.

\begin{figure}[H]
\centering
\fbox{%
  \begin{minipage}{0.98\linewidth}
  \raggedright
  \textbf{\textit{Role}}\\
  You are a professional text relevance evaluator with expertise in semantic analysis and content comparison.
  Your task is to objectively assess whether two texts are semantically equivalent, considering logic consistency, factuality, accuracy, and conciseness.\\

  \textbf{\textit{Task}}\\
  Determine whether the two texts are semantically equivalent.
  Return YES only if they convey essentially the same meaning, such as sharing the same core characters, objects, actions, scenes, and overall intent; otherwise return NO (unrelated, only weakly related, partially overlapping, or inconsistent). You should first briefly explain your reasoning, and then provide the final result.\\

  \textbf{\textit{Texts to compare}}\\
  (1). \{text\_1\}\\
  (2). \{text\_2\}\\
  
  \textbf{\textit{Output format}}\\
  \textbf{[Reason]:} Briefly explain the basis for your judgment, focusing on whether the two texts share the same topic, entities, actions, or overall meaning.\\
  \textbf{[Result]:} Yes / No
  \end{minipage}%
}
\captionsetup{justification=raggedright,singlelinecheck=false}
\caption{\textbf{LLM-as-a-judge prompt} used for OpenAI o3 verification.
}
\label{fig:judge_prompt}
\end{figure}
\definecolor{ablagray}{HTML}{F7F7F7}%
\definecolor{groupgray}{HTML}{F2F2F2}%
\begin{center}
\begin{minipage}{\textwidth}
\centering
\captionof{table}{\textbf{MCQ evaluation on VideoEspresso.}
We compare the VideoEspresso-trained ROVER-enhanced model with the Qwen2.5-VL-7B baseline.
\best{Best} results are highlighted.
}
\label{tab:videoespresso_mcq}
\setlength{\tabcolsep}{2.5pt}
\renewcommand{\arraystretch}{1.05}

\resizebox{\textwidth}{!}{%
\begin{tabular}{l|c|c|c|c|c|c|c|c|c|c|c|c|c|c|c}
\toprule
\textbf{Method}
& \textbf{Narra.} & \textbf{Event} & \textbf{Ingre.} & \textbf{Causal} & \textbf{Theme}
& \textbf{Conte.} & \textbf{Influ.} & \textbf{Role} & \textbf{Inter.} & \textbf{Behav.}
& \textbf{Emoti.} & \textbf{Cook.} & \textbf{Traff.} & \textbf{Situa.}
& \textbf{Avg.} \\
\midrule
\rowcolor{groupgray}
\multicolumn{16}{c}{\textit{Results on Qwen2.5-VL-7B}} \\
\midrule
Qwen2.5-VL
& 39.2 & 40.8 & 49.0 & 38.7 & 47.5
& 50.5 & \best{48.6} & 35.0 & 38.7 & \best{31.6}
& 41.5 & 45.3 & 43.3 & \best{58.0}
& 43.4 \\
+ $\mathrm{ROVER}$
& \best{51.9}  & \best{43.9} & \best{59.2} & \best{39.4} & \best{50.8}
& \best{54.1} & 47.2 & \best{47.6} & \best{43.5} & 29.8
& \best{44.6} & \best{47.2} & \best{56.7} & 52.0
& \best{47.7} \\
\bottomrule
\end{tabular}%
}
\end{minipage}
\end{center}

\subsection{Additional Qualitative Examples}
\label{sec:appendix_infercase}

\begin{figure}[t]
  \centering
  \begin{subfigure}{\textwidth}
    \centering
    \includegraphics[width=\textwidth]{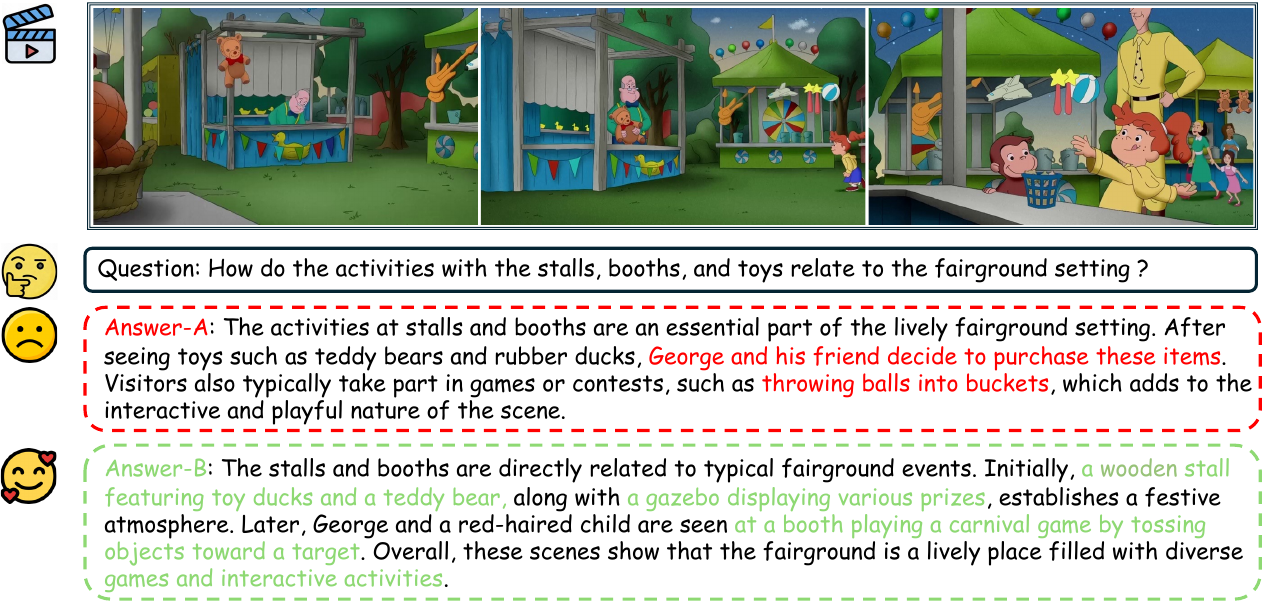}
    \caption{Example 1.}
    \label{fig:ve_infercase_diff_1}
  \end{subfigure}

  \begin{subfigure}{\textwidth}
    \centering
    \includegraphics[width=\textwidth]{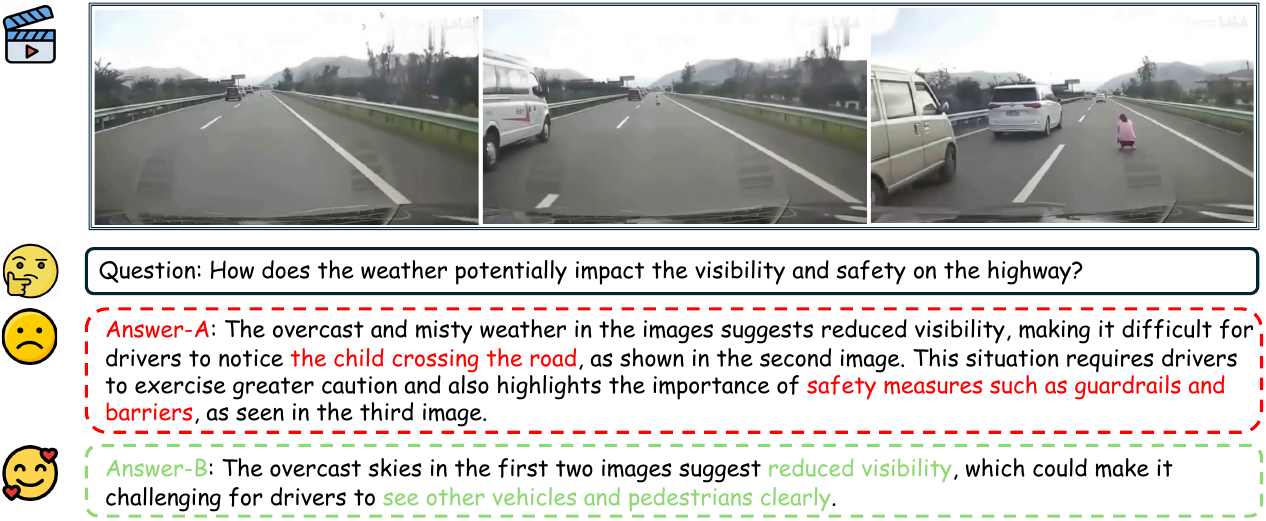}
    \caption{Example 2.}
    \label{fig:ve_infercase_diff_2}
  \end{subfigure}

  \begin{subfigure}{\textwidth}
    \centering
    \includegraphics[width=\textwidth]{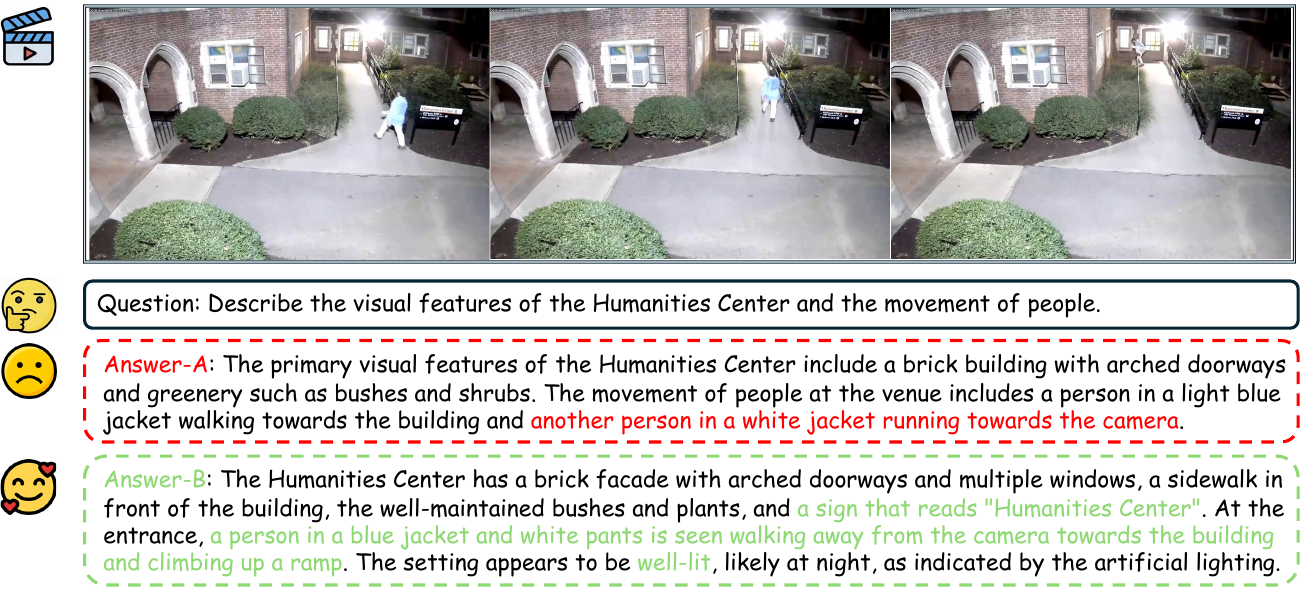}
    \caption{Example 3.}
    \label{fig:ve_infercase_diff_3}
  \end{subfigure}
  
    \caption{\textbf{Qualitative comparison on the VideoEspresso test set.} Compared to the RoI-resampling baseline (Answer-A), ROVER (Answer-B) demonstrates superior holistic scene understanding and precise reasoning of key inter-object relations, thereby yielding more accurate predictions.}
  \label{fig:ve_infercase_diff}
\end{figure}

\begin{figure}[t]
  \centering
  \begin{subfigure}{\textwidth}
    \centering
    \includegraphics[width=\textwidth]{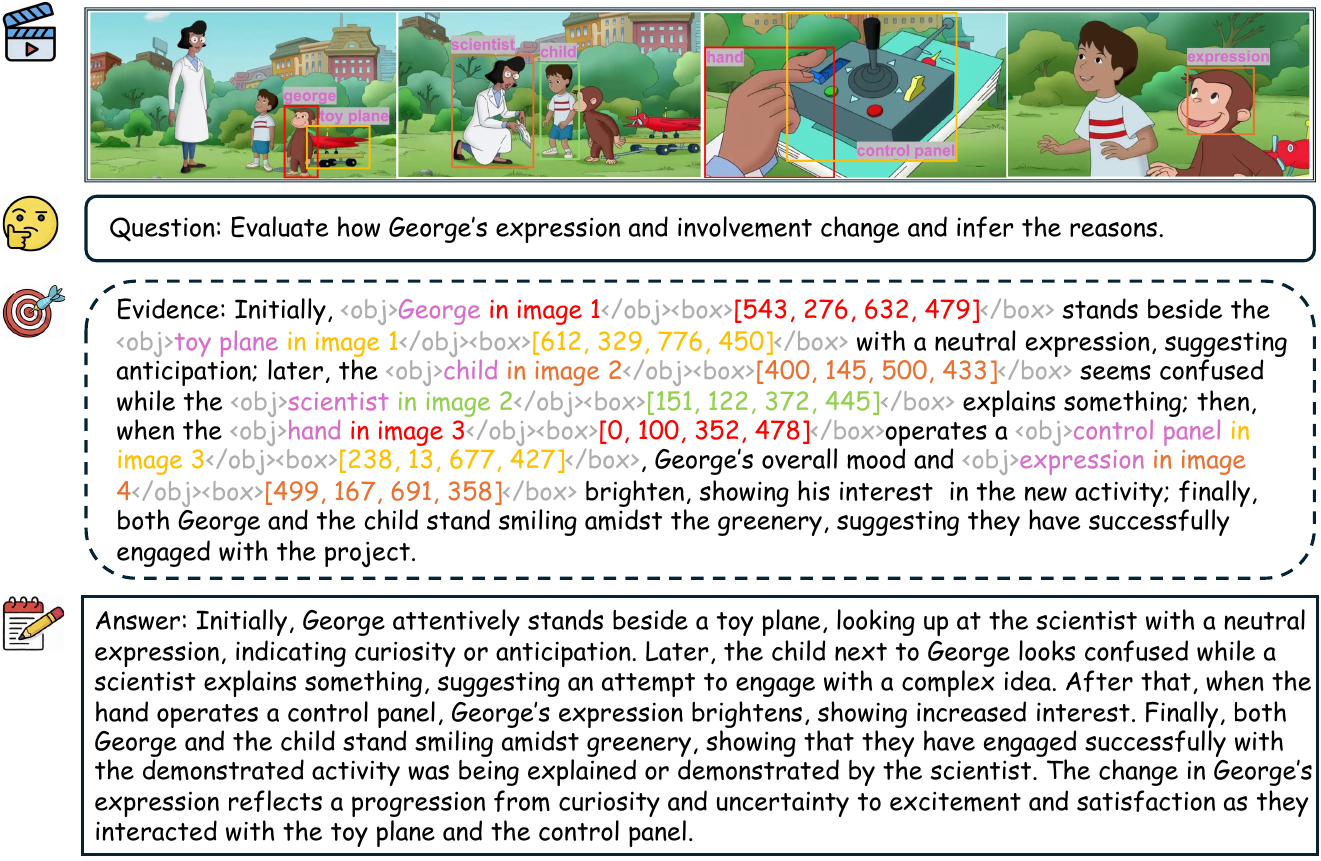}
    \caption{Example 1.}
    \label{fig:ve_infercase_1}
  \end{subfigure}

  \begin{subfigure}{\textwidth}
    \centering
    \includegraphics[width=\textwidth]{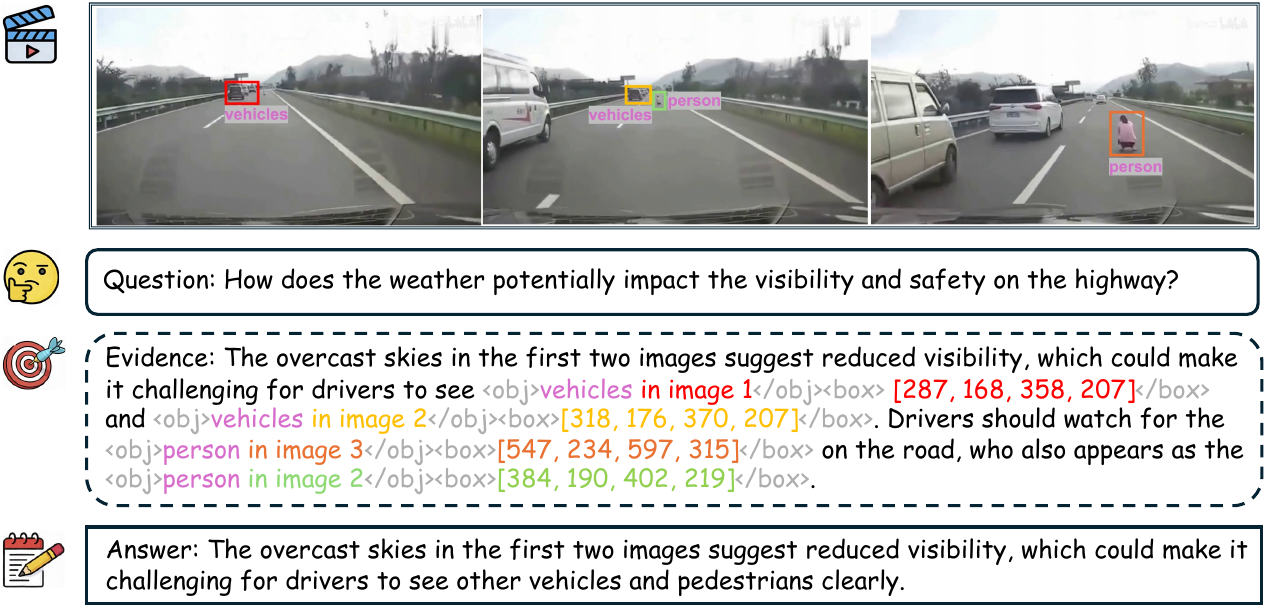}
    \caption{Example 2.}
    \label{fig:ve_infercase_2}
  \end{subfigure}

  \caption{\textbf{Additional qualitative examples on the VideoEspresso test set.} Model-predicted grounded evidence with bounding boxes across core frames, followed by the final answer.}
  \label{fig:ve_infercase}
\end{figure}

\begin{figure}[t]
  \centering
  \includegraphics[width=\textwidth]{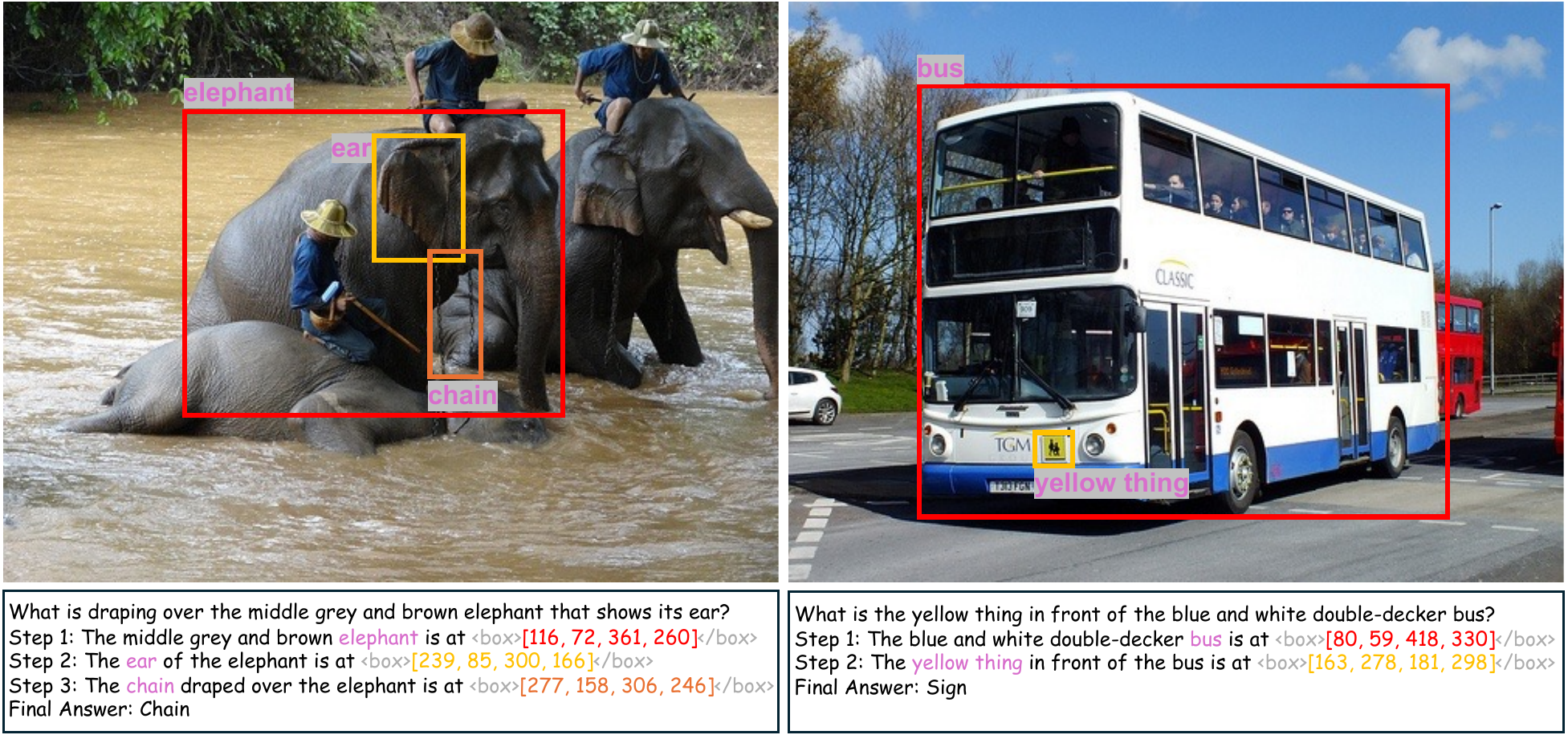}
  \caption{\textbf{Additional qualitative example on the MM-GCoT test set.} Model-predicted grounded evidence with bounding boxes, followed by the final answer.}
  \label{fig:mmgcot_infercase_1}
\end{figure}

\begin{figure}[t]
  \centering
  \includegraphics[width=\textwidth]{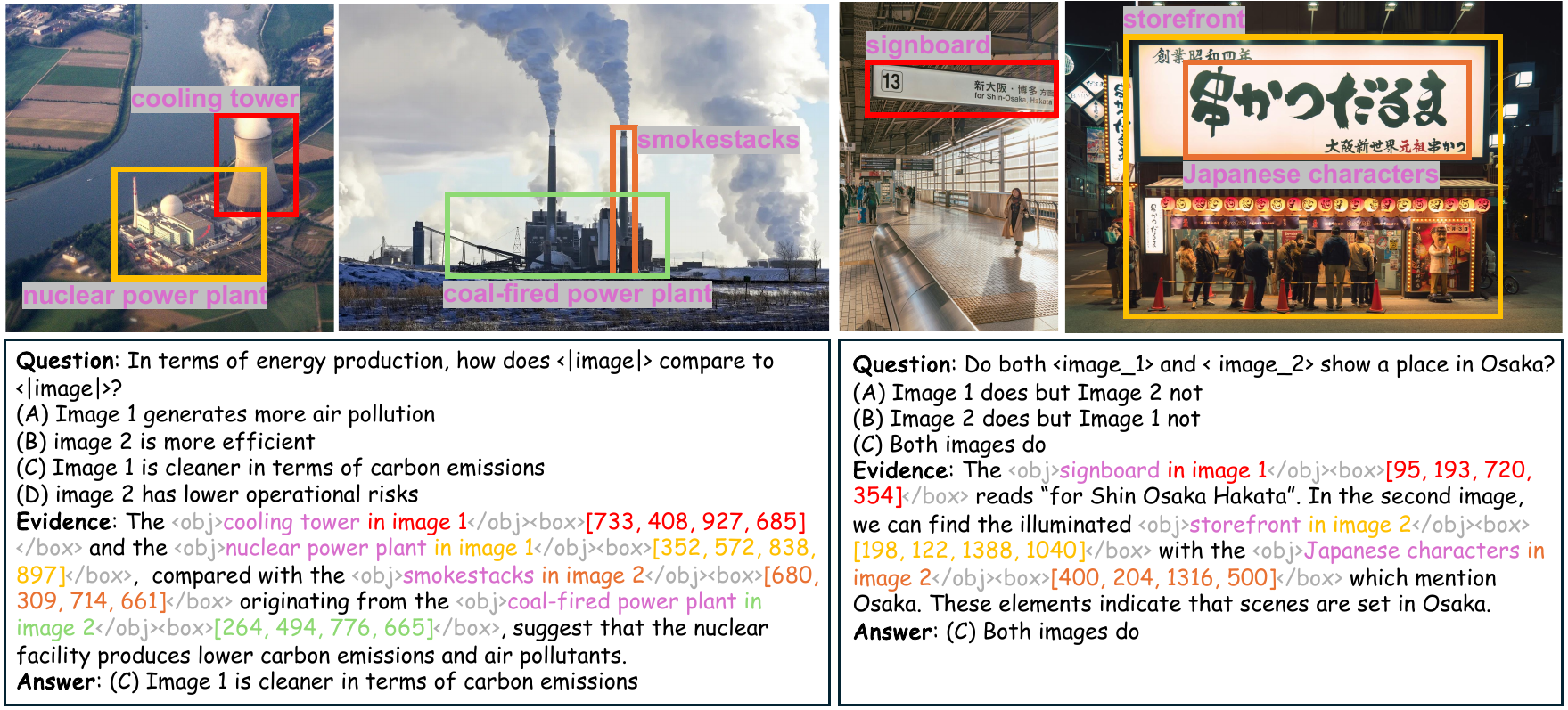}
  \caption{\textbf{Transfer example on Mantis after training on VideoEspresso.} Model-predicted multi-image reasoning that integrates cues from multiple objects and images to support the final answer.}
  \label{fig:mantis_infercase_1}
\end{figure}

\begin{figure}[t]
  \centering
  \includegraphics[width=\textwidth]{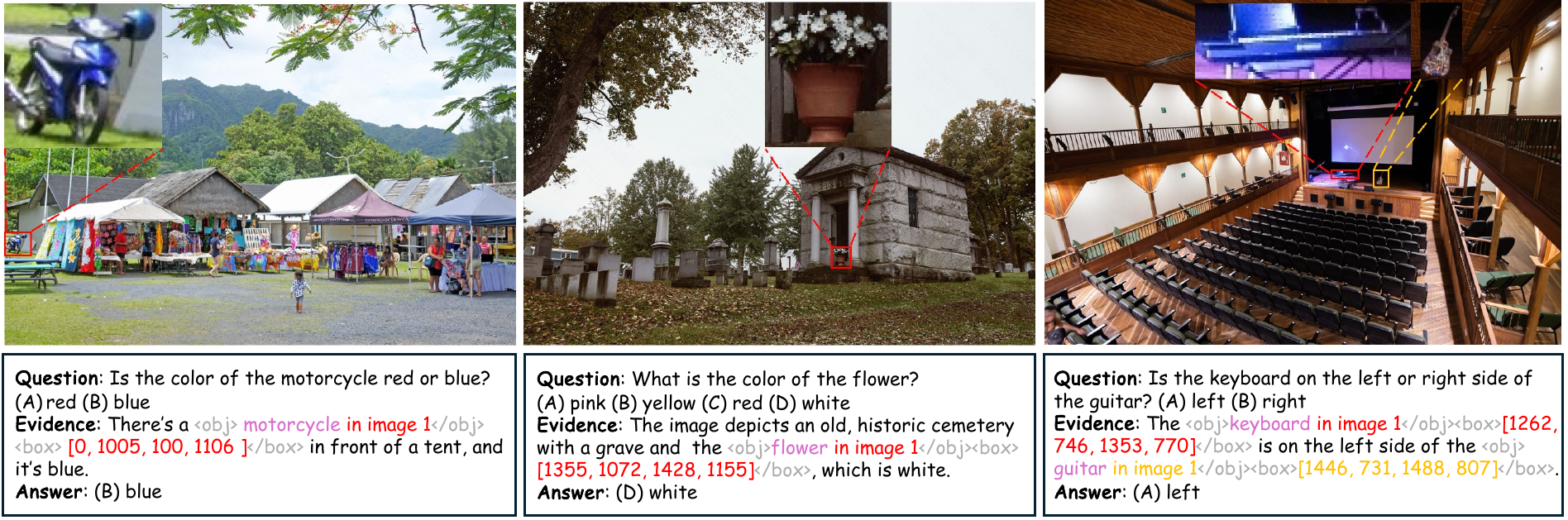}
  \caption{\textbf{Transfer example on V-Star after training on VideoEspresso.} Model-predicted high-resolution grounding and fine-grained recognition in a single-image setting.}
  \label{fig:vstar_infercase_1}
\end{figure}

\begin{figure}[t]
  \centering
  \includegraphics[width=\linewidth]{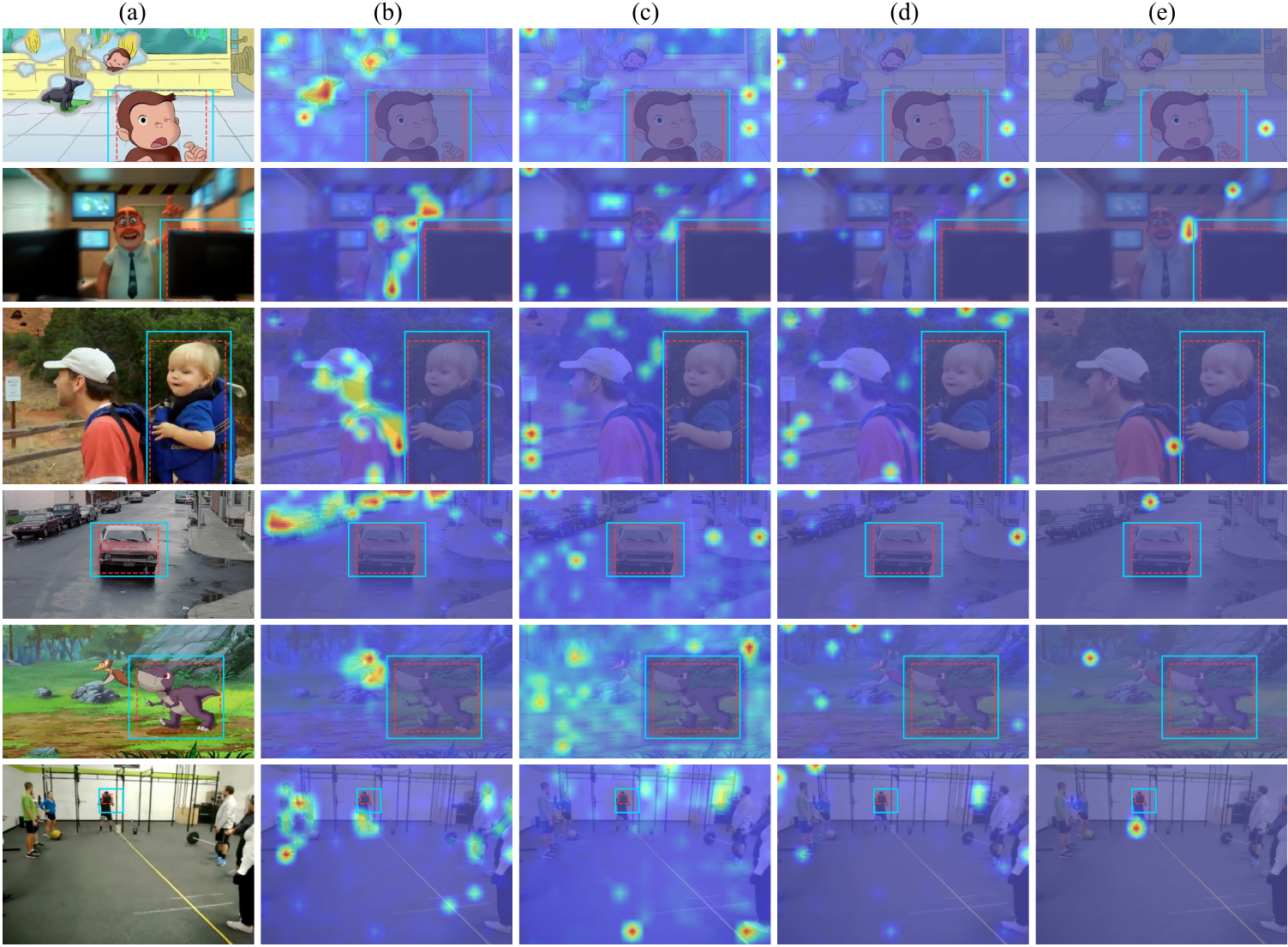}
    \caption{\textbf{Differential cross-attention visualization within \textbf{Sift}.}
      (a) Input image.
      (b--c) Positive and negative attention maps in DiffAttn, respectively.
      (d) Standard cross-attention replacing DiffAttn in \textbf{Sift}.
      (e) Cross-attention directly on raw ViT patch features.
      In (b--e), the shared query is the average-pooled feature over cyan-outlined patches overlapping the predicted box (red dashed).
      For clarity, each map is normalized to $[0,1]$ by its maximum value.}
  \label{fig:diffattn_vis}
\end{figure}

\begin{figure}[t]
  \centering
  \includegraphics[width=\linewidth]{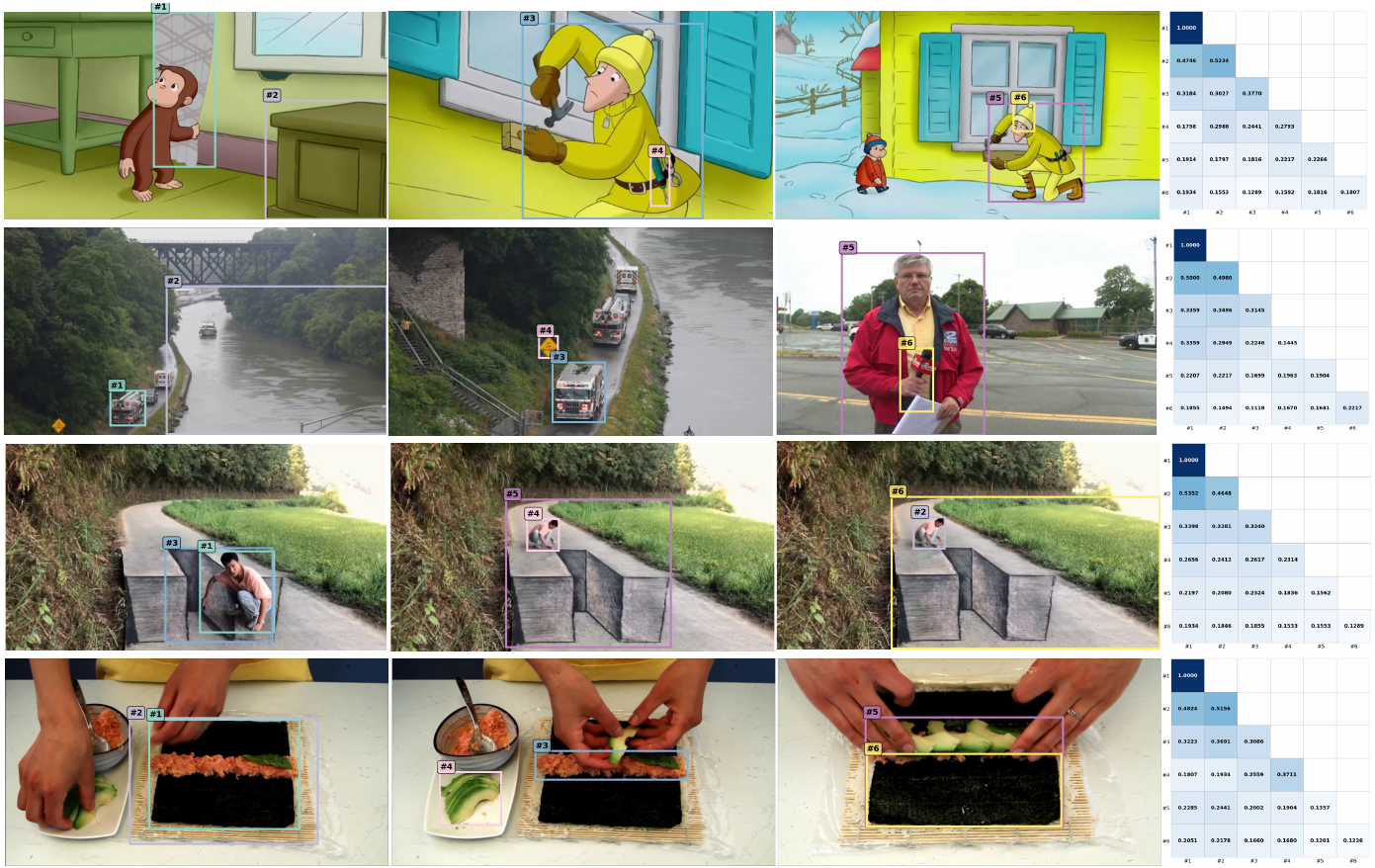}
  \caption{\textbf{VWS cross-attention visualization.} 
    We demonstrate \textbf{Weave} retrieving and aggregating relevant cues from previously routed objects into the current reasoning step.}
  \label{fig:memoattn_vis}
\end{figure}

To further illustrate the effectiveness and generalizability of our approach, we provide additional qualitative examples on both in-domain test sets (VideoEspresso: \Cref{fig:ve_infercase_diff} and \Cref{fig:ve_infercase}; MM-GCoT: \Cref{fig:mmgcot_infercase_1}) and diverse transfer benchmarks (Mantis: \Cref{fig:mantis_infercase_1}; V-Star: \Cref{fig:vstar_infercase_1}). For brevity, ROVER triplets following grounding predictions are omitted from the presented textual outputs. Moreover, to offer deeper insights into the model's inner workings, we include visualizations of both the differential cross-attention (\Cref{fig:diffattn_vis}) and the VWS cross-attention (\Cref{fig:memoattn_vis}).

\clearpage



\end{document}